\pdfoutput=1

\documentclass[11pt]{article}

\usepackage[]{acl}

\usepackage{times}
\usepackage{latexsym}

\usepackage[T1]{fontenc}

\usepackage[utf8]{inputenc}

\usepackage{microtype}

\usepackage{amsmath,amsfonts,bm}

\def\eqref#1{equation~\ref{#1}}

\def\1{\bm{1}}

\DeclareMathAlphabet{\mathsfit}{\encodingdefault}{\sfdefault}{m}{sl}
\SetMathAlphabet{\mathsfit}{bold}{\encodingdefault}{\sfdefault}{bx}{n}

\def\gC{{\mathcal{C}}}
\def\gD{{\mathcal{D}}}

\def\gL{{\mathcal{L}}}

\def\gU{{\mathcal{U}}}

\newcommand{\E}{\mathbb{E}}

\newcommand{\R}{\mathbb{R}}

\usepackage{graphicx}
\usepackage{footmisc}
\usepackage{caption}
\usepackage{subcaption}
\usepackage{booktabs}
\usepackage{xspace}
\usepackage{pifont}

\let\svthefootnote\thefootnote
\newcommand\blankfootnote[1]{%
  \let\thefootnote\relax\footnotetext{#1}%
  \let\thefootnote\svthefootnote%
}

\newcommand{\cmark}{\ding{51}}%
\newcommand{\xmark}{\ding{55}}%

\title{Sparse Interventions in Language Models with Differentiable Masking}

\author{Nicola De Cao~\textsuperscript{*1,2}, Leon Schmid~\textsuperscript{*3}, Dieuwke Hupkes~\textsuperscript{4}, Ivan Titov~\textsuperscript{1,2,5} \\
\textsuperscript{1}University of Amsterdam,
\textsuperscript{2}University of Edinburgh \\
\textsuperscript{3}University of Osnabrück,
\textsuperscript{4}Facebook AI Research,
\textsuperscript{5}Innopolis University\\
\href{mailto:nicola.decao@gmail.com}{\texttt{{\color{black} nicola.decao@gmail.com}}}, \
\href{mailto:lschmid@uos.de}{\texttt{{\color{black} lschmid@uos.de}}}\\
\href{mailto:dieuwkehupkes@fb.com}{\texttt{{\color{black} dieuwkehupkes@fb.com}}},\ \href{mailto:ititov@inf.ed.ac.uk}{\texttt{{\color{black} ititov@inf.ed.ac.uk}}}
}

\begin{document}
\maketitle
\begin{abstract}

There has been a lot of interest in understanding
what information is captured by hidden representations of language models (LMs). %
Typically, interpretation methods i) do not guarantee that the model actually uses the encoded information, and ii) do not discover small subsets of neurons responsible for a considered phenomenon. 
Inspired by causal mediation analysis, we propose a method that discovers within a neural LM a small subset of neurons responsible for a particular linguistic phenomenon, i.e., subsets causing a change
in the corresponding token emission probabilities.
We use a differentiable relaxation to approximately search through the combinatorial space. An $L_0$ regularization term ensures that the
search converges to discrete and sparse solutions.    %
We apply our method to analyze subject-verb number agreement and gender bias detection in LSTMs. We observe that it is fast and finds better solutions than the alternative (REINFORCE). Our experiments confirm that each of these phenomenons is mediated through a small subset of neurons that do not play any other discernible role. 

\end{abstract}

\blankfootnote{\textsuperscript{*}Equal contributions.}

\section{Introduction}

The success of language models (LMs) in many natural language processing tasks is accompanied by an increasing interest in interpreting and analyzing such models. %
One goal in this direction is to identify how a model employs its hidden representations to arrive at a prediction~\citep{belinkov2019analysis,jacovi-goldberg-2020-towards}.
A popular line of research studies LMs with ``diagnostic classifiers'' or ``probes'' that are trained to predict linguistics properties from hidden units, with the purpose of analyzing what information is encoded by the network and where~\citep{alain2016understanding,adi2016fine,hupkes2018visualisation,voita-titov-2020-information}.
However, this method is sometimes criticized for generating unfaithful interpretations~\citep{barrett-etal-2019-adversarial} since the trained classifiers only measure the correlation between a model's representations and an external property and not whether such property is actually causing the model's predictions. Indeed, several studies pointed out limitations of probes \citep{belinkov2019analysis, vanmassenhove2017investigating,tamkin-etal-2020-investigating}, including mismatches between the performance of the probe and the original model as well as the discrepancy between correlation and causation of hidden units and model outputs.

In response to these limitations, several recent studies have proposed to interpret neural models with \textit{interventions} which aim to measure causal effects by intervening in representations of the model and observing a change in the model output \cite{giulianelli-etal-2018-hood,elazar2021amnesic,feder2021causalm}.
These techniques let us investigate more directly if an LM represents a certain linguistic phenomenon but are limited when it comes to understanding where and how this information is represented.
Therefore, an important question that they cannot answer is to what extent \emph{modularity} -- often believed to be a prerequisite for systematic generalization \cite{goyal2020inductive,dittadi2020transfer} -- is a property that emerges naturally in such models.
An adaptation of \textit{causal mediation analysis}~\citep{pearl2001direct} by~\citet{lakretz-etal-2019-emergence,vig2020investigating,lakretz2021mechanisms} makes an important step towards enabling such investigations. 
They consider neurons one by one by setting their activation to zero and measuring their effect on the output. 
However,  
these techniques suffer from two major shortcomings:
i) as systematically ablating combinations of multiple neurons is computationally infeasible, they are restricted to detecting single neurons ii) there is no guarantee that setting a unit activation to zero corresponds to switching the corresponding function on or off~\citep{sturmfels2020visualizing}.

In this work, we use a differentiable relaxation of this search problem to overcome both these limitations. More specifically, our goal is 
to identify neurons responsible for shifting the probability from a word to its alternative in examples exemplifying the phenomena, without affecting other LM predictions.  For example, when investigating subject-verb number agreement, we want to redistribute the probability mass
from the singular form of an upcoming verb to the plural one (or the other way around), while discouraging changes in the distributions for other contexts. In this way, we ensure that the function is mediated through the detected neurons, and these neurons do not play any other discernible role.

Building on the framework of differentiable masking~\citep{de-cao-etal-2020-decisions,schlichtkrull2021interpreting}, we formalize this search for a {\it sparse intervention} as a constrained optimization problem. We aim to both detect the responsible neurons and learn the values to assign them when intervening. We use a continuous relaxation of the subset-selection problem, but
ensure discreteness and encourage sparsity through
 $L_0$ regularization. The $L_0$ penalty determines how many neurons we want to discover. 
In our experiments, we use an LSTM-based LM, previously investigated by ~\citep{gulordava-etal-2018-colorless,lakretz-etal-2019-emergence}, and consider subject-verb number agreement and gender bias detection. We start with validating our method by showing that we can replicate findings reported in these previous studies and then dive into a deeper analysis. We show that our proposed method is effective as well as computationally efficient (i.e., it converges up to 7 times faster than REINFORCE).

\section{Additional Related Work}

The $L_0$ regularization was first proposed by~\citet{louizos2018learning} in the context of pruning. It has been used in a variety of works in NLP as a tool for generating rationals and  attribution~\citep{bastings-etal-2019-interpretable,de-cao-etal-2020-decisions,schlichtkrull2021interpreting}. Masking weights and groups of weights was also used by~\citet{csordas2020neural} to investigate the functional modularity of neural networks. 

A number of studies suggested that some of the linguistic phenomena are encoded, at least to a large degree, in a disentangled and sparse fashion. For example,  \citet{radford2017learning} detected a neuron encoding sentiment polarity and \citet{dai2021knowledge} showed that individual facts learned by an LM can be manipulated by modifying a small number of neurons. In a similar spirit, \citet{voita-etal-2019-analyzing} observed that many Transformer attention heads in a neural machine translation model are specialized; interestingly, they also used $L_0$ regularization but only to prune less important heads; the roles played by the heads were identified by studying their attention patterns. Our technique can facilitate such studies by effectively identifying sets of neural network's subcomponents playing a given function.

\section{Method}
A range of tests for causal language models consider if a model can represent a particular linguistic phenomenon \citep[i.e., subject-verb-agreement, filler gap dependencies, negative polarity items][]{jumelet-etal-2021-language,jumelet-etal-2019-analysing,wilcox-etal-2018-rnn,gulordava-etal-2018-colorless}, by measuring whether that model assigns a higher probability to a grammatical sentence involving that phenomenon than to its minimally different ungrammatical counterpart.
In such tests, the comparison of probabilities is often focused on the probability of a single token -- for instance, the probability of the correct and incorrect verb-form in a long sentence \citep{linzen-etal-2016-assessing}.
Here, we exploit this paradigm and investigate if we can find groups of neurons for which a modification of their value -- 
which we call an \emph{intervention} -- systematically leads to a change of probability for the single token emission related to a specific phenomenon.

Because there is no direct supervision for interventions (i.e., we do not have a dataset that accompanies a model with annotated examples of interventions), we need to learn them with a proxy objective.
Let's assume we have an autoregressive model (e.g., an LSTM;~\citealt{HochSchm97}) that assigns a probability to sequences.
For a set of input tokens $x = \langle x_1, \ldots, x_n \rangle$, we obtain the model's probability of the token of interest $p(x_n|x_{<n})$ along with the hidden states $h = \langle h_1, \ldots, h_n \rangle$ where $h_i \in \R^k$ (one for each time step). We then intervene in the model's computation by modifying a group of neurons from one or multiple hidden states.
The intervention at a certain time step $i\!<\!n$ consists of a binary mask $m \in \{0,1\}^k$ indicating which hidden units need intervention and which can be left unchanged. The intervention is then made substituting the $i$th hidden state with the altered state 
\begin{equation}
    \hat h_i = (1 - m) \odot h_i + m \odot b \;,
\end{equation}
where $\odot$ indicates the element-wise product and $b \in \R^k$ is a learned baseline vector that will lead the desired intervention. 
We denote $p(x_n|x_{<n},\hat h_i)$ as the model's probability of the target token when its forward pass has been altered using $\hat h_i$.

In addition, as the main objective of this work, we are looking for sparse interventions, which we define as finding a  defined small percentage (e.g., 1-5\%) of neurons where to apply an intervention to while keeping all the rest untouched.

\subsection{Learning to Intervene} \label{sec:method_learning}
Because there is no direct supervision to estimate the mask $m$ and the baseline $b$, we train them to minimize the ratio
\begin{equation}
    \gL_{\text{ratio}}(\hat h_i,x) = \frac{p(x_n=d|x_{<n},\hat h_i)}{p(x_n=t|x_{<n},\hat h_i)} \;,
\end{equation}
where we want to identify neurons responsible for a change in probability between a predicted word $d$ and a target word $t$ (e.g., a correct and incorrect verb form---where it does not matter what the model predicts but we set $d$ as the form for which the model assigns higher probability and $t$ as the other). 
In other words, we optimize them to assign more probability mass to the token $t$ rather than $d$. In addition, we desire such intervention to be as sparse as possible because we want to identify the least number of neurons responsible for the decision. Such \textit{sparsity} corresponds to constraining most of the entries of $m$ to be $0$, which corresponds to not interfering. We cast this %
in the language of constrained optimization.

A practical way to express the sparsity constraint is through the $L_0$ `norm'.\footnote{$L_0$, denoted $\|m \|_0$ and defined as $\#(i|m_i \neq 0)$, is the number of non-zeros entries in a vector. Contrary to $L_1$ or $L_2$, $L_0$ is not a homogeneous function and, thus, not a proper norm. However, contemporary literature refers to it as a norm, and we do so as well to avoid confusion.} Our constraint is defined as the total number of neurons we intervene on:
\begin{equation} \label{eq:l0}
    \gC_0(m) = \sum_{i=1}^k \mathbf{1}_{[\R_{\neq 0}]}(m_i) \;.
\end{equation}
The whole optimization problem is then:
\begin{align}
    \min_{m,b} &\quad \sum_{x \in \gD} \gL_{\text{ratio}}(\hat h_i,x) \\
    \mathrm{s.t.} &\quad \gC_0(m) \leq \alpha \;,
\end{align}
where $\gD$ is a dataset and the margin $\alpha \in (0,1]$ is a hyperparameter that controls the desired sparsity we want to achieve (i.e., the lower the sparser the solution will be). Since non-linear constrained optimization is generally intractable, we employ Lagrangian relaxation~\citep{boyd2004convex} optimizing instead
\begin{equation} \label{eq:lagrangian_objective}
	\max_\lambda \min_{m_i,b} \; \sum_{x \in \gD} \gL_{\text{ratio}}(\hat h_i) + \lambda(\gC_0(m_i) - \alpha) \;,
\end{equation}
where $\lambda \in \R_{\geq 0}$ is the Lagrangian multiplier. 
Since we use binary masks, our loss is discontinuous and non-differentiable. A default option would be to use REINFORCE~\citep{williams1992simple}, but it is known to have a noisy gradient and thus slow convergence. Then, to overcome both problems, we resort to a sparse relaxation to binary variables, namely using a Hard Concrete distribution~\citep{louizos2018learning}. See Appendix~\ref{app:relaxation} for further details and proprieties of that distribution.

\subsection{Single-step and Every-step intervention}
We described how we apply an intervention at a certain time step $i\!<\!n$ as an intervention that directly modifies $h_i$. Thus, we refer to this type as a \textit{single-step} intervention. The choice of the time step to intervene is not arbitrary and should be carefully set to investigate a particular phenomenon in the LM---note that this is task-dependent, e.g., for exploring subject-verb number agreement, a reasonable choice is to make the intervention at the hidden state of the subject.
As an extension, we also define an \textit{every-step} intervention when instead of altering only $h_i$ we modify all $h_1,\ldots,h_{n-1}$ with the same $m$ and $b$ (not that this is the same type of intervention used by~\citealt{lakretz-etal-2019-emergence}).
The two types of intervention investigate different proprieties of an LM, and we experiment with both variants.

\subsection{Retaining other predictions}
We train interventions to modify the model's prediction at a specific token position. However, there is little guarantee that all the other token positions will have the same output distribution as without the interventions.
This is important as, when investigating modularity, we would like to ensure not only that a group of neurons plays a distinct interpretable role but also that they do not fulfil any other discernable role. 
For this reason, we employ a regularization term in addition to the constrained objective. This corresponds to minimizing a KL divergence between the output distributions of the original model and the one from the model with interventions (see Appendix~\ref{app:regularization} for more details).

\section{Experimental Setting} \label{sec:experiments}

We study the pre-trained LSTM language model made available by~\citet{gulordava-etal-2018-colorless}\footnote{Model and pre-trained corpus available from the authors at \url{https://github.com/facebookresearch/colorlessgreenRNNs}}, which has been studied extensively in previous work and therefore provides a good testing ground for our method.
We study the model, as well as newly trained models with the same architecture, on two tasks described below: subject-verb number agreement and gender bias. 
The evaluation for tasks naturally follows the defined objective $\gL_{\text{ratio}}(\hat h_i,x)$ (see Section~\ref{sec:method_learning}). Without intervention, the ratio is always $>1$, and we defined a successful intervention when we find a mask and baseline values such that the ratio becomes $<1$. 
Finally, we define the \textit{accuracy} of interventions as the average of times when the ratio is $<1$ across all datapoints in a given dataset/task. In other words, the accuracy reflects how often we can \emph{flip} the model's decision (that corresponds to making the likelihood of desired output higher than the previous model output).

\paragraph{LSTM language model}
The studied model is a standard two-layered LSTM with a hidden dimension of 650. The embedding layer also has dimensionality 650, and it is not tied with the output layer. The vocabulary size is 50,000 (the most frequent words in the corpus). The model was trained on English Wikipedia data (with around 80M tokens training tokens and 10M for validation). We used this model to compare to previous findings of~\citet{lakretz-etal-2019-emergence}. We also pre-train this LM several times with different weights initializations to make sure our results generalize. 

\paragraph{Subject-verb number agreement task}
This task consists of intervening on the subject-verb number agreement predicted from the model: for a given sentence, we wish the intervention to change the number of the verb from singular to plural or the other way around. 
We employ data made available by~\citet{lakretz-etal-2019-emergence}\footnote{\url{https://github.com/FAIRNS/Number_and_syntax_units_in_LSTM_LMs}} (see Appendix~\ref{app:data} for more details). We apply the single-step intervention to the subject of the (only) verb. We apply two types of intervention: turning the verb to the singular form or the plural one.

\paragraph{Gender bias detection}
This task consists of intervening on a pronoun that might indicate a gender bias of the model: for a given sentence, we wish the intervention to change the pronoun that refers to a person with a profession and an unspecified gender. 
We employ data made available by~\citet{vig2020investigating}\footnote{\url{https://github.com/sebastianGehrmann/CausalMediationAnalysis}} (see Appendix~\ref{app:data} for more details). Also for his task, we apply the single-step intervention to the subject of the (only) verb. We apply two types of intervention one for flipping the pronoun to ``\textit{he}'' and one for ``\textit{she}''.

\section{Results}

\paragraph{Main results}
For the single-step intervention (with regularization), our method achieved $88.8 \pm0.3$ and $95.8 \pm3.8$ accuracies for the number agreement and gender bias tasks, respectively. On average, our method finds $5$ and $5.7$ units for the two tasks, respectively.  Considering that the LM has 1300 hidden units, this intervention is relatively sparse as desired (i.e., we use $<0.44\%$ of the total units).

\begin{figure}[t]
    \centering
    \includegraphics[width=0.48\textwidth]{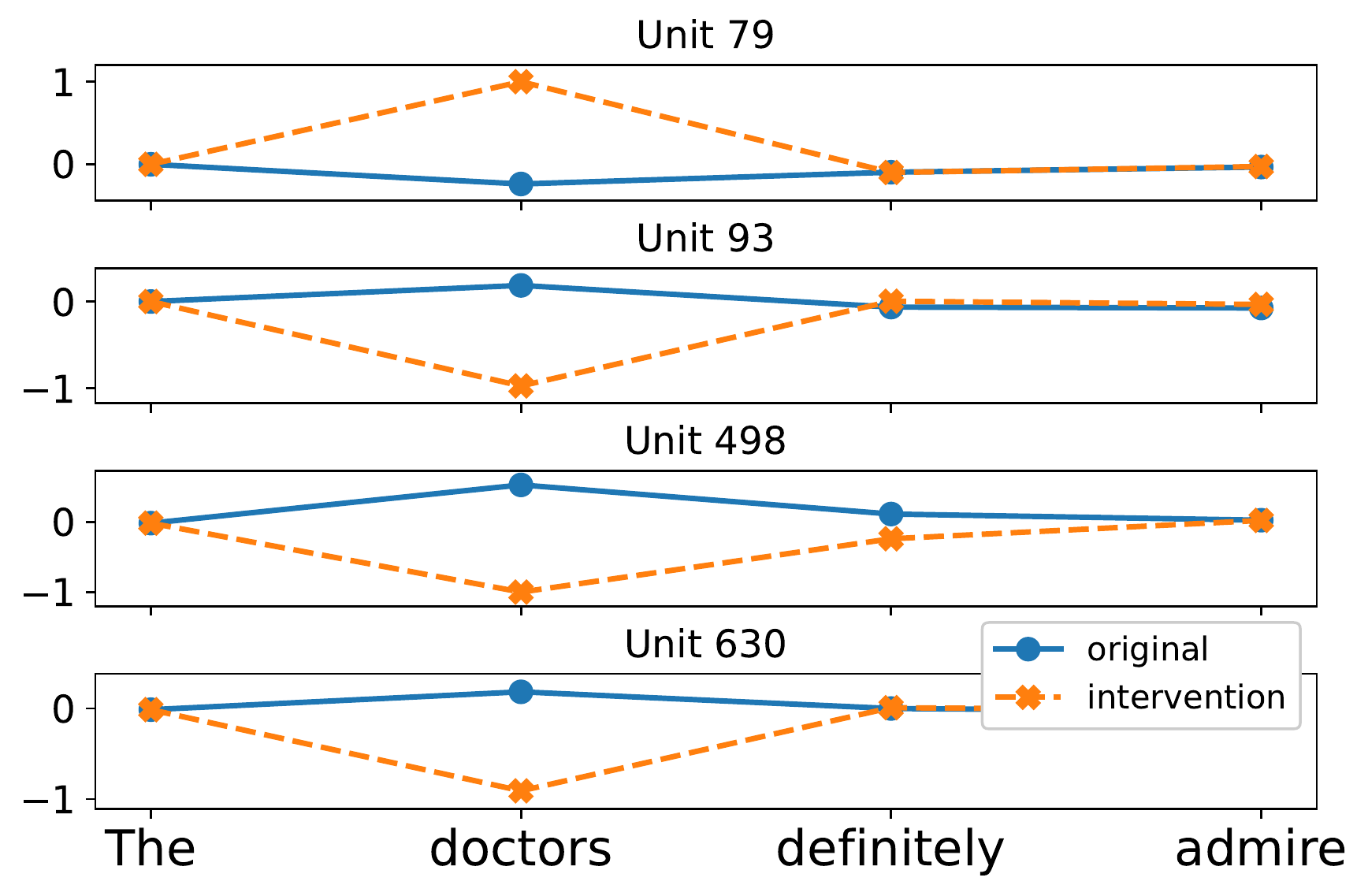}
    \caption{Activations of four units we intervene on (single step intervention at ``doctors'') for changing number agreement (at ``admire'').}
    \label{fig:example_na}
\end{figure}

\begin{figure}[t]
    \centering
    \includegraphics[width=0.48\textwidth]{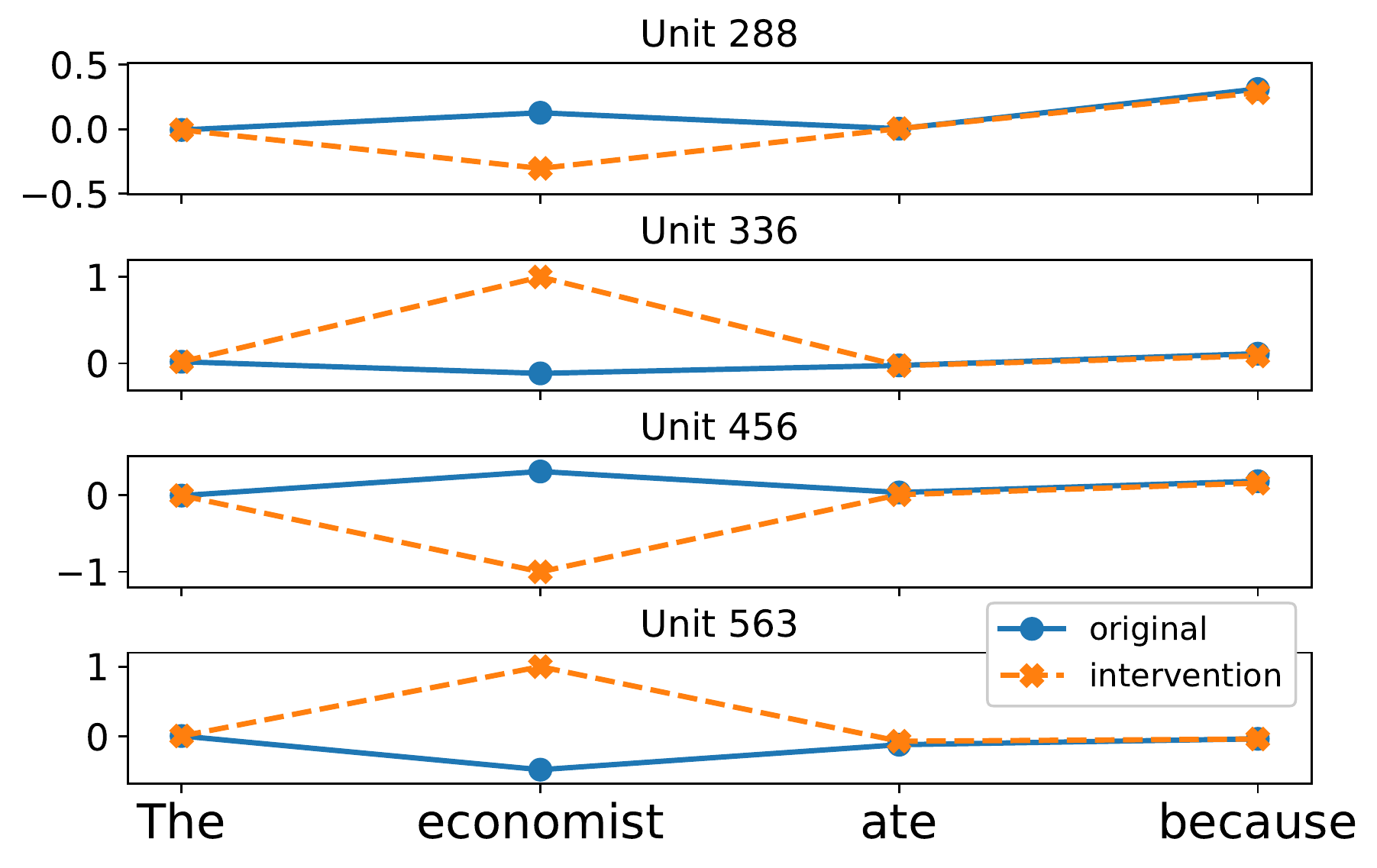}
    \caption{Activations of four units we intervene on (single step intervention at ``economist'') for changing the pronoun (after ``because'').}
    \label{fig:example_gb}
\end{figure}

In Figure~\ref{fig:example_na} and~\ref{fig:example_gb}, we show examples of hidden state activations with and without the intervention for both tasks (see Appendix~\ref{app:additional_results} for additional examples). From these figures, we can see that only one unit is heavily affected (the one for the intervention) while the others are minimally corrupted after that time step. We hypothesize that the model stores the information of number or gender in other units (or in cell states), but the discovered units are the ones responsible for the \textit{initialization} of such memory. In Table~\ref{tab:svna_single} and~\ref{tab:gb_single} in Appendix~\ref{app:additional_results} we report the the full list of discovered units and the learned baseline vectors for both tasks on the single-step intervention.

For the every-step intervention, our method achieved an almost perfect accuracy of $99.3$ and $100$ for the number agreement and gender bias tasks, respectively, while using $3.3$ units on average for both tasks. This type of intervention is much more effective and more intrusive (i.e., the number of changed units is larger as it happens at every step). In Table~\ref{tab:svna_every} we report the full list of discovered units and the learned baseline vectors comparing to the one discovered by~\citet{lakretz-etal-2019-emergence} (every-step intervention). Noticeably, we \textit{re-discover} units 776 and 988 which validates our method and confirm their findings. Interestingly, we also discover an extra unit on average, highlighting that one of the limitations of~\citet{lakretz-etal-2019-emergence} was indeed an efficient way to search units. For a summary of all results see Table~\ref{tab:summary}, and for the discovered units and baseline on the gender task see Table~\ref{tab:gb_every} (in Appendix~\ref{app:additional_results}).

\begin{table}[t]
    \centering
    \begin{tabular}{lrrr}
    \toprule
    & \textbf{Accuracy} & \textbf{Units} & \textbf{KL} \\ 
    \midrule
    \midrule
    & \multicolumn{3}{c}{Number agreement} \\
    \midrule
    Single & 88.0 \tiny{$\pm 0.2$} & 5.0 \tiny{$\pm 0.0$} & 0.034 \tiny{$\pm 0.003$} \\
    Single\textsuperscript{R} & 88.8 \tiny{$\pm 0.3$} & 5.0 \tiny{$\pm 0.0$} & 0.032 \tiny{$\pm 0.000$}\\
    \midrule
    Every & 99.8 \tiny{$\pm 0.2$} & 3.3 \tiny{$\pm 0.6$} & 0.208 \tiny{$\pm 0.023$} \\
    Every\textsuperscript{R} & 99.3 \tiny{$\pm 0.2$}& 3.3 \tiny{$\pm 0.6$} & 0.075 \tiny{$\pm 0.005$}\\
    \midrule
    \midrule
    & \multicolumn{3}{c}{Gender bias} \\
    \midrule
    Single & 98.6 \tiny{$\pm 1.2$} & 7.0 \tiny{$\pm 1.0 $} & 0.011 \tiny{$\pm 0.001$}\\
    Single\textsuperscript{R} & 95.8 \tiny{$\pm 3.8$}& 5.7 \tiny{$\pm 1.2$} & 0.009 \tiny{$\pm 0.002$} \\
    \midrule
    Every & 100.0 \tiny{$\pm 0.0$}& 2.7 \tiny{$\pm 0.6$} & 0.104 \tiny{$\pm 0.020$}\\
    Every\textsuperscript{R} & 100.0 \tiny{$\pm 0.0$}& 3.3 \tiny{$\pm 0.0$} & 0.075 \tiny{$\pm 0.001$} \\
    \bottomrule
    \end{tabular}
    \caption{Summary of results for both the number agreement and gender bias settings (average across 3 run for each setting). \textsuperscript{R} indicates KL regularization. Single/ Every indicates single-step and every-step interventions respectively.}
    \label{tab:summary}
\end{table}

\begin{table*}[t]
    \centering
    \begin{tabular}{rrrrc}
    \toprule
    \textbf{Unit} & \textbf{Singular} & \textbf{Plural} & \textbf{Prevalence} & \textbf{Found by}~\citet{lakretz-etal-2019-emergence} \\
    \midrule
    79 & -0.05 \tiny{$\pm 0.11$} & 0.53 \tiny{$\pm 0.13$} & 100\% & \xmark \\
    224 & -0.67 \tiny{$\pm 0.03$} & 0.48 \tiny{$\pm 0.06$} & 32\% & \xmark \\
    498 & 0.53 \tiny{$\pm 0.10$} & -0.32 \tiny{$\pm 0.03$} & 100\% & \xmark \\
    630 & 0.43 \tiny{$\pm 0.06$} & -0.36 \tiny{$\pm 0.08$} & 56\% & \xmark \\
    776 & -1.00 \tiny{$\pm 0.00$} & 0.52 \tiny{$\pm 0.03$} & 100\% & \cmark \\
    988 & 0.19 \tiny{$\pm 0.06$} & -1.00 \tiny{$\pm 0.00$} & 44\% & \cmark \\
    \bottomrule
    \end{tabular}
    \caption{Subject-verb number agreement task with every-step interventions. Values are averages across 25 runs. ``Found'' indicates how many times our model decides to apply the intervention on a specific unit across runs.}
    \label{tab:svna_every}
\end{table*}

\begin{table}[t]
    \centering
    \resizebox*{0.48\textwidth}{!}{  
    \fontsize{8.4}{10.1}\selectfont
    \begin{tabular}{lrrr}
    \toprule
    & Acc. ($\uparrow$) & Units ($\downarrow$) & Speed ($\downarrow$) \\
    \midrule
    SFE ($\alpha=0.05$)    & \textbf{100.0}  & 20.0          & 5.2h \\
    SFE ($\alpha=0.02$)    & 87.6          & 6.0           & 3.6h \\
    \textbf{Ours} ($\alpha=0.02$)      & 92.3          & \textbf{4.3}  & \textbf{0.5h} \\
    \bottomrule
    \end{tabular}
    }
    \caption{Comparison between the solutions find by Score Function Estimation (aka REINFORCE) and our system (average across 25 runs). Ours is much faster and if finds a sparser solution with better accuracy.}
    \label{tab:reinfoce_vs_ours}
\end{table}

\paragraph{Efficiency}
To demonstrate the efficiency and efficacy of our estimation employing the Hard Concrete distribution, we compared to the standard Score Function Estimation (aka REINFORCE;~\citealt{williams1992simple}) with a moving average baseline for variance reduction. Table~\ref{tab:reinfoce_vs_ours} summarises the results for the single-step intervention. We tried different values for $\alpha$ for REINFORCE to achieve a reasonable trade-off between accuracy and the number of units used. REINFORCE takes at least 7 times more time to converge, and it always converges at using more units than our method with lower accuracy. 
We did not compare to an exact search (i.e., complete evaluate all the combinations of units and optimizing the intervention baselines) since it scales exponentially with the number of units---it has a time complexity of $O(2^k)$ where $k$ is the number of neurons. A full search would correspond to $>10^{12}$ evaluations only for checking combinations up to 4 neurons which is clearly unfeasible. 

\paragraph{Robustness}
To demonstrate that our method is robust, we tested it on 5 language models initialized with different seeds and trained with the original script by~\citet{gulordava-etal-2018-colorless}. We run our method for the single-step intervention 3 times for each language model. The average accuracy at convergence is $88.7\pm 2.6$, and we discover $4.7\pm 0.5$ units on average. The variability in both accuracy and number of units is very low, indicating that our method is robust to different parameterizations of the model we applied it to.

\paragraph{Effect of Regularization}
We ablated the KL regularization to see whether it affects learning and the final convergence of our method. On the number agreement task, we found that the average KL divergence with respect to the original model predictions was $0.009/0.075$ with regularization and $0.011/0.104$ without regularization (for single-step and every-step intervention, respectively). We used different regularization coefficients (i.e., different weights), but we did not observe a substantial change in the convergence of our models. Moreover, the accuracy and the number of units found with regularization was almost the same as without regularization (see Table~\ref{tab:summary} for all results). This lack of effect of the regularization suggests the studied phenomenon is naturally captured by specialized neurons.
In the gender bias task, regularization has a slightly higher impact. The regularized method converges to finding fewer units on average and with worse accuracy ($95.8$ as opposed to $98.6$) in the single-step intervention. In the every-step intervention, the accuracy stays invariant (for both settings is $100$) while the model converges to using more units.

\section{Conclusions}
In this work, we present a new method that employs constraint optimization to efficiently find hidden units that are responsible for particular language phenomena of language models. We additionally use an $L_0$ regularization to find a sparse solution which we also show empirically--i.e., our methodology discovers few units in the order of $5$-$6$ that is $<0.44\%$ of all units in the studied LM. We show such sparse solutions can be found for multiple phenomena (number agreement and gender) and is an useful tool for analysis of what a LM has learned and how units influence its token emissions.
Although this works focuses on LSTM models, the proposed technique is not architecture-dependent and thus easily applicable to transformer-based models as well as convolution-based model and many others.

\clearpage
\subsection*{Acknowledgements}
NDC and IT are supported by SAP Innovation Center Network; LS and IT by the Dutch Organization for Scientific Research (NWO VIDI 639.022.518) and IT by the Analytical Center for the Government of Russian Federation (agreement 70-2021-00143, dd. 01.11.2021, IGK 000000D730321P5Q0002).

\bibliography{anthology,custom}

\begin{thebibliography}{41}
\expandafter\ifx\csname natexlab\endcsname\relax\def\natexlab#1{#1}\fi

\bibitem[{Adi et~al.(2017)Adi, Kermany, Belinkov, Lavi, and
  Goldberg}]{adi2016fine}
Yossi Adi, Einat Kermany, Yonatan Belinkov, Ofer Lavi, and Yoav Goldberg. 2017.
\newblock \href {https://openreview.net/forum?id=BJh6Ztuxl} {Fine-grained
  analysis of sentence embeddings using auxiliary prediction tasks}.
\newblock In \emph{5th International Conference on Learning Representations,
  {ICLR} 2017, Toulon, France, April 24-26, 2017, Conference Track
  Proceedings}. OpenReview.net.

\bibitem[{Alain and Bengio(2017)}]{alain2016understanding}
Guillaume Alain and Yoshua Bengio. 2017.
\newblock \href {https://openreview.net/pdf?id=HJ4-rAVtl} {Understanding
  intermediate layers using linear classifier probes}.
\newblock \emph{International Conference on Learning Representations}.

\bibitem[{Barrett et~al.(2019)Barrett, Kementchedjhieva, Elazar, Elliott, and
  S{\o}gaard}]{barrett-etal-2019-adversarial}
Maria Barrett, Yova Kementchedjhieva, Yanai Elazar, Desmond Elliott, and Anders
  S{\o}gaard. 2019.
\newblock \href {https://doi.org/10.18653/v1/D19-1662} {Adversarial removal of
  demographic attributes revisited}.
\newblock In \emph{Proceedings of the 2019 Conference on Empirical Methods in
  Natural Language Processing and the 9th International Joint Conference on
  Natural Language Processing (EMNLP-IJCNLP)}, pages 6330--6335, Hong Kong,
  China. Association for Computational Linguistics.

\bibitem[{Bastings et~al.(2019)Bastings, Aziz, and
  Titov}]{bastings-etal-2019-interpretable}
Jasmijn Bastings, Wilker Aziz, and Ivan Titov. 2019.
\newblock \href {https://doi.org/10.18653/v1/P19-1284} {Interpretable neural
  predictions with differentiable binary variables}.
\newblock In \emph{Proceedings of the 57th Annual Meeting of the Association
  for Computational Linguistics}, pages 2963--2977, Florence, Italy.
  Association for Computational Linguistics.

\bibitem[{Belinkov and Glass(2019)}]{belinkov2019analysis}
Yonatan Belinkov and James Glass. 2019.
\newblock \href {https://doi.org/10.1162/tacl_a_00254} {Analysis methods in
  neural language processing: A survey}.
\newblock \emph{Transactions of the Association for Computational Linguistics},
  7:49--72.

\bibitem[{Bolukbasi et~al.(2016)Bolukbasi, Chang, Zou, Saligrama, and
  Kalai}]{bolukbasi2016man}
Tolga Bolukbasi, Kai{-}Wei Chang, James~Y. Zou, Venkatesh Saligrama, and
  Adam~Tauman Kalai. 2016.
\newblock \href
  {https://proceedings.neurips.cc/paper/2016/hash/a486cd07e4ac3d270571622f4f316ec5-Abstract.html}
  {Man is to computer programmer as woman is to homemaker? debiasing word
  embeddings}.
\newblock In \emph{Advances in Neural Information Processing Systems 29: Annual
  Conference on Neural Information Processing Systems 2016, December 5-10,
  2016, Barcelona, Spain}, pages 4349--4357.

\bibitem[{Boyd et~al.(2004)Boyd, Boyd, and Vandenberghe}]{boyd2004convex}
Stephen Boyd, Stephen~P Boyd, and Lieven Vandenberghe. 2004.
\newblock \href {https://doi.org/10.1017/CBO9780511804441} {\emph{Convex
  optimization}}.
\newblock Cambridge university press.

\bibitem[{Csord{\'{a}}s et~al.(2021)Csord{\'{a}}s, van Steenkiste, and
  Schmidhuber}]{csordas2020neural}
R{\'{o}}bert Csord{\'{a}}s, Sjoerd van Steenkiste, and J{\"{u}}rgen
  Schmidhuber. 2021.
\newblock \href {https://openreview.net/forum?id=7uVcpu-gMD} {Are neural nets
  modular? inspecting functional modularity through differentiable weight
  masks}.
\newblock In \emph{9th International Conference on Learning Representations,
  {ICLR} 2021, Virtual Event, Austria, May 3-7, 2021}. OpenReview.net.

\bibitem[{Dai et~al.(2021)Dai, Dong, Hao, Sui, and Wei}]{dai2021knowledge}
Damai Dai, Li~Dong, Yaru Hao, Zhifang Sui, and Furu Wei. 2021.
\newblock \href {https://arxiv.org/abs/2104.08696} {Knowledge neurons in
  pretrained transformers}.
\newblock \emph{ArXiv preprint}, abs/2104.08696.

\bibitem[{De~Cao et~al.(2020)De~Cao, Schlichtkrull, Aziz, and
  Titov}]{de-cao-etal-2020-decisions}
Nicola De~Cao, Michael~Sejr Schlichtkrull, Wilker Aziz, and Ivan Titov. 2020.
\newblock \href {https://doi.org/10.18653/v1/2020.emnlp-main.262} {How do
  decisions emerge across layers in neural models? interpretation with
  differentiable masking}.
\newblock In \emph{Proceedings of the 2020 Conference on Empirical Methods in
  Natural Language Processing (EMNLP)}, pages 3243--3255, Online. Association
  for Computational Linguistics.

\bibitem[{Dittadi et~al.(2021)Dittadi, Tr{\"{a}}uble, Locatello, Wuthrich,
  Agrawal, Winther, Bauer, and Sch{\"{o}}lkopf}]{dittadi2020transfer}
Andrea Dittadi, Frederik Tr{\"{a}}uble, Francesco Locatello, Manuel Wuthrich,
  Vaibhav Agrawal, Ole Winther, Stefan Bauer, and Bernhard Sch{\"{o}}lkopf.
  2021.
\newblock \href {https://openreview.net/forum?id=8VXvj1QNRl1} {On the transfer
  of disentangled representations in realistic settings}.
\newblock In \emph{9th International Conference on Learning Representations,
  {ICLR} 2021, Virtual Event, Austria, May 3-7, 2021}. OpenReview.net.

\bibitem[{Elazar et~al.(2021)Elazar, Ravfogel, Jacovi, and
  Goldberg}]{elazar2021amnesic}
Yanai Elazar, Shauli Ravfogel, Alon Jacovi, and Yoav Goldberg. 2021.
\newblock \href {https://doi.org/10.1162/tacl_a_00359} {{Amnesic Probing:
  Behavioral Explanation with Amnesic Counterfactuals}}.
\newblock \emph{Transactions of the Association for Computational Linguistics},
  9:160--175.

\bibitem[{Feder et~al.(2021)Feder, Oved, Shalit, and
  Reichart}]{feder2021causalm}
Amir Feder, Nadav Oved, Uri Shalit, and Roi Reichart. 2021.
\newblock \href {https://doi.org/10.1162/coli_a_00404} {{CausaLM: Causal Model
  Explanation Through Counterfactual Language Models}}.
\newblock \emph{Computational Linguistics}, 47(2):333--386.

\bibitem[{Giulianelli et~al.(2018)Giulianelli, Harding, Mohnert, Hupkes, and
  Zuidema}]{giulianelli-etal-2018-hood}
Mario Giulianelli, Jack Harding, Florian Mohnert, Dieuwke Hupkes, and Willem
  Zuidema. 2018.
\newblock \href {https://doi.org/10.18653/v1/W18-5426} {Under the hood: Using
  diagnostic classifiers to investigate and improve how language models track
  agreement information}.
\newblock In \emph{Proceedings of the 2018 {EMNLP} Workshop {B}lackbox{NLP}:
  Analyzing and Interpreting Neural Networks for {NLP}}, pages 240--248,
  Brussels, Belgium. Association for Computational Linguistics.

\bibitem[{Goyal and Bengio(2020)}]{goyal2020inductive}
Anirudh Goyal and Yoshua Bengio. 2020.
\newblock \href {https://arxiv.org/abs/2011.15091} {Inductive biases for deep
  learning of higher-level cognition}.
\newblock \emph{ArXiv preprint}, abs/2011.15091.

\bibitem[{Gulordava et~al.(2018)Gulordava, Bojanowski, Grave, Linzen, and
  Baroni}]{gulordava-etal-2018-colorless}
Kristina Gulordava, Piotr Bojanowski, Edouard Grave, Tal Linzen, and Marco
  Baroni. 2018.
\newblock \href {https://doi.org/10.18653/v1/N18-1108} {Colorless green
  recurrent networks dream hierarchically}.
\newblock In \emph{Proceedings of the 2018 Conference of the North {A}merican
  Chapter of the Association for Computational Linguistics: Human Language
  Technologies, Volume 1 (Long Papers)}, pages 1195--1205, New Orleans,
  Louisiana. Association for Computational Linguistics.

\bibitem[{Hochreiter and Schmidhuber(1997)}]{HochSchm97}
Sepp Hochreiter and Jürgen Schmidhuber. 1997.
\newblock \href {http://dx.doi.org/10.1162/neco.1997.9.8.1735} {Long short-term
  memory}.
\newblock \emph{Neural Computation}, 9(8):1735--1780.

\bibitem[{Hupkes et~al.(2018)Hupkes, Veldhoen, and
  Zuidema}]{hupkes2018visualisation}
Dieuwke Hupkes, Sara Veldhoen, and Willem Zuidema. 2018.
\newblock \href {https://www.ijcai.org/proceedings/2018/796} {Visualisation and
  'diagnostic classifiers' reveal how recurrent and recursive neural networks
  process hierarchical structure}.
\newblock \emph{Journal of Artificial Intelligence Research}, 61:907--926.

\bibitem[{Jacovi and Goldberg(2020)}]{jacovi-goldberg-2020-towards}
Alon Jacovi and Yoav Goldberg. 2020.
\newblock \href {https://doi.org/10.18653/v1/2020.acl-main.386} {Towards
  faithfully interpretable {NLP} systems: How should we define and evaluate
  faithfulness?}
\newblock In \emph{Proceedings of the 58th Annual Meeting of the Association
  for Computational Linguistics}, pages 4198--4205, Online. Association for
  Computational Linguistics.

\bibitem[{Jang et~al.(2017)Jang, Gu, and Poole}]{jang2016categorical}
Eric Jang, Shixiang Gu, and Ben Poole. 2017.
\newblock \href {https://openreview.net/forum?id=rkE3y85ee} {Categorical
  reparameterization with gumbel-softmax}.
\newblock In \emph{5th International Conference on Learning Representations,
  {ICLR} 2017, Toulon, France, April 24-26, 2017, Conference Track
  Proceedings}. OpenReview.net.

\bibitem[{Jumelet et~al.(2021)Jumelet, Denic, Szymanik, Hupkes, and
  Steinert-Threlkeld}]{jumelet-etal-2021-language}
Jaap Jumelet, Milica Denic, Jakub Szymanik, Dieuwke Hupkes, and Shane
  Steinert-Threlkeld. 2021.
\newblock \href {https://doi.org/10.18653/v1/2021.findings-acl.439} {Language
  models use monotonicity to assess {NPI} licensing}.
\newblock In \emph{Findings of the Association for Computational Linguistics:
  ACL-IJCNLP 2021}, pages 4958--4969, Online. Association for Computational
  Linguistics.

\bibitem[{Jumelet et~al.(2019)Jumelet, Zuidema, and
  Hupkes}]{jumelet-etal-2019-analysing}
Jaap Jumelet, Willem Zuidema, and Dieuwke Hupkes. 2019.
\newblock \href {https://doi.org/10.18653/v1/K19-1001} {Analysing neural
  language models: Contextual decomposition reveals default reasoning in number
  and gender assignment}.
\newblock In \emph{Proceedings of the 23rd Conference on Computational Natural
  Language Learning (CoNLL)}, pages 1--11, Hong Kong, China. Association for
  Computational Linguistics.

\bibitem[{Kingma and Welling(2014)}]{kingma2014auto}
Diederik~P. Kingma and Max Welling. 2014.
\newblock \href {http://arxiv.org/abs/1312.6114} {Auto-encoding variational
  bayes}.
\newblock In \emph{2nd International Conference on Learning Representations,
  {ICLR} 2014, Banff, AB, Canada, April 14-16, 2014, Conference Track
  Proceedings}.

\bibitem[{Lakretz et~al.(2021)Lakretz, Hupkes, Vergallito, Marelli, Baroni, and
  Dehaene}]{lakretz2021mechanisms}
Yair Lakretz, Dieuwke Hupkes, Alessandra Vergallito, Marco Marelli, Marco
  Baroni, and Stanislas Dehaene. 2021.
\newblock \href
  {https://doi.org/https://doi.org/10.1016/j.cognition.2021.104699} {Mechanisms
  for handling nested dependencies in neural-network language models and
  humans}.
\newblock \emph{Cognition}, 213:104699.
\newblock Special Issue in Honour of Jacques Mehler, Cognition’s founding
  editor.

\bibitem[{Lakretz et~al.(2019)Lakretz, Kruszewski, Desbordes, Hupkes, Dehaene,
  and Baroni}]{lakretz-etal-2019-emergence}
Yair Lakretz, German Kruszewski, Theo Desbordes, Dieuwke Hupkes, Stanislas
  Dehaene, and Marco Baroni. 2019.
\newblock \href {https://doi.org/10.18653/v1/N19-1002} {The emergence of number
  and syntax units in {LSTM} language models}.
\newblock In \emph{Proceedings of the 2019 Conference of the North {A}merican
  Chapter of the Association for Computational Linguistics: Human Language
  Technologies, Volume 1 (Long and Short Papers)}, pages 11--20, Minneapolis,
  Minnesota. Association for Computational Linguistics.

\bibitem[{Linzen et~al.(2016)Linzen, Dupoux, and
  Goldberg}]{linzen-etal-2016-assessing}
Tal Linzen, Emmanuel Dupoux, and Yoav Goldberg. 2016.
\newblock \href {https://doi.org/10.1162/tacl_a_00115} {Assessing the ability
  of {LSTM}s to learn syntax-sensitive dependencies}.
\newblock \emph{Transactions of the Association for Computational Linguistics},
  4:521--535.

\bibitem[{Louizos et~al.(2018)Louizos, Welling, and
  Kingma}]{louizos2018learning}
Christos Louizos, Max Welling, and Diederik~P. Kingma. 2018.
\newblock \href {https://openreview.net/forum?id=H1Y8hhg0b} {Learning sparse
  neural networks through l{\_}0 regularization}.
\newblock In \emph{6th International Conference on Learning Representations,
  {ICLR} 2018, Vancouver, BC, Canada, April 30 - May 3, 2018, Conference Track
  Proceedings}. OpenReview.net.

\bibitem[{Lu et~al.(2020)Lu, Mardziel, Wu, Amancharla, and
  Datta}]{lu2020gender}
Kaiji Lu, Piotr Mardziel, Fangjing Wu, Preetam Amancharla, and Anupam Datta.
  2020.
\newblock {Gender Bias in Neural Natural Language Processing}.
\newblock \emph{Logic, Language, and Security}, pages 189--202.

\bibitem[{Maddison et~al.(2017)Maddison, Mnih, and Teh}]{maddison2016concrete}
Chris~J. Maddison, Andriy Mnih, and Yee~Whye Teh. 2017.
\newblock \href {https://openreview.net/forum?id=S1jE5L5gl} {The concrete
  distribution: {A} continuous relaxation of discrete random variables}.
\newblock In \emph{5th International Conference on Learning Representations,
  {ICLR} 2017, Toulon, France, April 24-26, 2017, Conference Track
  Proceedings}. OpenReview.net.

\bibitem[{Pearl(2001)}]{pearl2001direct}
Judea Pearl. 2001.
\newblock {Direct and Indirect Effects}.
\newblock \emph{Uncertainty in Artificial Intelligence}, page 411–420.

\bibitem[{Radford et~al.(2017)Radford, Jozefowicz, and
  Sutskever}]{radford2017learning}
Alec Radford, Rafal Jozefowicz, and Ilya Sutskever. 2017.
\newblock \href {https://arxiv.org/abs/1704.01444} {Learning to generate
  reviews and discovering sentiment}.
\newblock \emph{ArXiv preprint}, abs/1704.01444.

\bibitem[{Rezende et~al.(2014)Rezende, Mohamed, and
  Wierstra}]{rezende2014stochastic}
Danilo~Jimenez Rezende, Shakir Mohamed, and Daan Wierstra. 2014.
\newblock \href {http://proceedings.mlr.press/v32/rezende14.html} {Stochastic
  backpropagation and approximate inference in deep generative models}.
\newblock In \emph{Proceedings of the 31th International Conference on Machine
  Learning, {ICML} 2014, Beijing, China, 21-26 June 2014}, volume~32 of
  \emph{{JMLR} Workshop and Conference Proceedings}, pages 1278--1286.
  JMLR.org.

\bibitem[{Schlichtkrull et~al.(2021)Schlichtkrull, Cao, and
  Titov}]{schlichtkrull2021interpreting}
Michael~Sejr Schlichtkrull, Nicola~De Cao, and Ivan Titov. 2021.
\newblock \href {https://openreview.net/forum?id=WznmQa42ZAx} {Interpreting
  graph neural networks for {NLP} with differentiable edge masking}.
\newblock In \emph{9th International Conference on Learning Representations,
  {ICLR} 2021, Virtual Event, Austria, May 3-7, 2021}. OpenReview.net.

\bibitem[{Sturmfels et~al.(2020)Sturmfels, Lundberg, and
  Lee}]{sturmfels2020visualizing}
Pascal Sturmfels, Scott Lundberg, and Su-In Lee. 2020.
\newblock \href {https://distill.pub/2020/attribution-baselines} {{Visualizing
  the Impact of Feature Attribution Baselines }}.
\newblock \emph{Distill}, 5(1):e22.

\bibitem[{Tamkin et~al.(2020)Tamkin, Singh, Giovanardi, and
  Goodman}]{tamkin-etal-2020-investigating}
Alex Tamkin, Trisha Singh, Davide Giovanardi, and Noah Goodman. 2020.
\newblock \href {https://doi.org/10.18653/v1/2020.findings-emnlp.125}
  {Investigating transferability in pretrained language models}.
\newblock In \emph{Findings of the Association for Computational Linguistics:
  EMNLP 2020}, pages 1393--1401, Online. Association for Computational
  Linguistics.

\bibitem[{Vanmassenhove et~al.(2017)Vanmassenhove, Du, and
  Way}]{vanmassenhove2017investigating}
Eva Vanmassenhove, Jinhua Du, and Andy Way. 2017.
\newblock {Investigating ‘Aspect’in NMT and SMT: Translating the English
  simple past and present perfect}.
\newblock \emph{Computational Linguistics in the Netherlands Journal},
  7:109--128.

\bibitem[{Vig et~al.(2020)Vig, Gehrmann, Belinkov, Qian, Nevo, Singer, and
  Shieber}]{vig2020investigating}
Jesse Vig, Sebastian Gehrmann, Yonatan Belinkov, Sharon Qian, Daniel Nevo,
  Yaron Singer, and Stuart~M. Shieber. 2020.
\newblock \href
  {https://proceedings.neurips.cc/paper/2020/hash/92650b2e92217715fe312e6fa7b90d82-Abstract.html}
  {Investigating gender bias in language models using causal mediation
  analysis}.
\newblock In \emph{Advances in Neural Information Processing Systems 33: Annual
  Conference on Neural Information Processing Systems 2020, NeurIPS 2020,
  December 6-12, 2020, virtual}.

\bibitem[{Voita et~al.(2019)Voita, Talbot, Moiseev, Sennrich, and
  Titov}]{voita-etal-2019-analyzing}
Elena Voita, David Talbot, Fedor Moiseev, Rico Sennrich, and Ivan Titov. 2019.
\newblock \href {https://doi.org/10.18653/v1/P19-1580} {Analyzing multi-head
  self-attention: Specialized heads do the heavy lifting, the rest can be
  pruned}.
\newblock In \emph{Proceedings of the 57th Annual Meeting of the Association
  for Computational Linguistics}, pages 5797--5808, Florence, Italy.
  Association for Computational Linguistics.

\bibitem[{Voita and Titov(2020)}]{voita-titov-2020-information}
Elena Voita and Ivan Titov. 2020.
\newblock \href {https://doi.org/10.18653/v1/2020.emnlp-main.14}
  {Information-theoretic probing with minimum description length}.
\newblock In \emph{Proceedings of the 2020 Conference on Empirical Methods in
  Natural Language Processing (EMNLP)}, pages 183--196, Online. Association for
  Computational Linguistics.

\bibitem[{Wilcox et~al.(2018)Wilcox, Levy, Morita, and
  Futrell}]{wilcox-etal-2018-rnn}
Ethan Wilcox, Roger Levy, Takashi Morita, and Richard Futrell. 2018.
\newblock \href {https://doi.org/10.18653/v1/W18-5423} {What do {RNN} language
  models learn about filler{--}gap dependencies?}
\newblock In \emph{Proceedings of the 2018 {EMNLP} Workshop {B}lackbox{NLP}:
  Analyzing and Interpreting Neural Networks for {NLP}}, pages 211--221,
  Brussels, Belgium. Association for Computational Linguistics.

\bibitem[{Williams(1992)}]{williams1992simple}
Ronald~J. Williams. 1992.
\newblock \href {https://doi.org/10.1007/BF00992696} {Simple statistical
  gradient-following algorithms for connectionist reinforcement learning}.
\newblock \emph{Mach. Learn.}, 8(3–4):229–256.

\end{thebibliography}
\bibliographystyle{acl_natbib}

\clearpage
\appendix
\section{Stochastic relaxation of the Mask} \label{app:relaxation}

Our optimization problem poses two difficulties: i) $\gC_0$ is discontinuous and has zero derivative almost everywhere, and ii) the altered state $\hat h_i$ is discontinuous w.r.t. the binary mask $m$. A simple way to overcome both issues is to treat the binary mask as stochastic and optimize the objective in expectation.
In that case, one natural option is to resort to score function estimation~\citep[REINFORCE;][]{williams1992simple} while another is to use a sparse relaxation to binary variables~\cite{louizos2018learning,bastings-etal-2019-interpretable,de-cao-etal-2020-decisions,schlichtkrull2021interpreting}. In Section~\ref{sec:experiments} we discuss the two aforementioned options showing that the latter is much more effective (results in Table~\ref{tab:reinfoce_vs_ours}).
Thus we opt to use the Hard Concrete distribution, a mixed discrete-continuous distribution on the closed interval $[0, 1]$. This distribution assigns a non-zero probability to exactly zero and one while it also admits continuous outcomes in the unit interval via the \textit{reparameterization trick}~\citep{kingma2014auto}. We refer to~\citet{louizos2018learning} for details, but also provide a brief summary in Appendix~\ref{sec:binary_concrete}. With a stochastic mask, the objective is computed in expectation, which addresses both sources of non-differentiability. Note that during training the mask is sampled and its values lies in the closed unit interval. After training, we set the mask entries to exact ones when their expected values are $>0.5$ or to zero otherwise.
To prevent issues due to the discrepancy between the values of the mask during training and during inference, we add another constraint 
\begin{equation}
    \gC_{(0,1)} = \sum_{i=1}^k \E[m_i \in (0,1)] \;,
\end{equation}
to be $\leq \beta$. $\gC_{(0,1)}$ during training constrains the relaxed mask values not to lie in the open interval $(0, 1)$ but rather to concentrate in $\{0,1\}$. $\beta \in (0,1]$ is an hyperparameter (the lower the less discrepancy is expected).

\section{The Hard Concrete distribution} \label{sec:binary_concrete}

The Hard Concrete distribution, assigns density to continuous outcomes in the open interval $(0,1)$ and non-zero mass to exactly $0$ and exactly $1$. A particularly appealing property of this distribution is that sampling can be done via a differentiable reparameterization~\citep{rezende2014stochastic, kingma2014auto}. 
In this way, the $\gC_0$ constrain in Equation~\ref{eq:l0} becomes an expectation
\begin{equation} \label{eq:l0_relaxation}
	\gC_0(m) = \sum_{i=1}^k \E_{p(m_i)} \left[ m_i \neq 0 \right] \;.
\end{equation}
whose gradient can be estimated via Monte Carlo sampling without the need for REINFORCE and without introducing biases. We did modify the original Hard Concrete, though only so slightly, in a way that it gives support to samples in the half-open interval $[0, 1)$, that is,  with non-zero mass only at $0$. That is because we need only distinguish $0$ from non-zero, and the value $1$ is not particularly important.\footnote{Only a true $0$ is guaranteed to completely mask an input out, while any non-zero value, however small, may leak some amount of information.} 

\paragraph{The distribution}
A stretched and rectified Binary Concrete (also known as Hard Concrete) distribution is obtained applying an affine transformation to the Binary Concrete distribution~\citep{maddison2016concrete, jang2016categorical} and rectifying its samples in the interval $[0,1]$. A Binary Concrete is defined over the open interval $(0, 1)$ and it is parameterised by a location parameter $\gamma \in \R$ and temperature parameter $\tau \in \R_{>0}$. The location acts as a logit and it controls the probability mass skewing the distribution towards $0$ in case of negative location and towards $1$ in case of positive location. The temperature parameter controls the concentration of the distribution. The Binary Concrete is then stretched with an affine transformation extending its support to $(l, r)$ with $l \leq 0$ and $r \geq 1$. Finally, we obtain a Hard Concrete distribution rectifying samples in the interval $[0,1]$. This corresponds to collapsing the probability mass over the interval $(l, 0]$ to $0$, and the mass over the interval $[1, r)$ to $1$. This induces a distribution over the close interval $[0, 1]$ with non-zero mass at $0$ and $1$. Samples are obtained using
\begin{equation}
	\begin{aligned}
		s & = \sigma \left( \left(\log u - \log(1 - u) + \gamma  \right) / \tau  \right)   \;, \\
		z & = \min \left( 1, \max \left( 0, s \cdot \left( l - r \right) + r \right) \right) \;, 
	\end{aligned}
\end{equation}
where $\sigma$ is the Sigmoid function $\sigma(x) = (1 + e^{-x})^{-1}$ and $u\sim\gU(0,1)$. We point to the Appendix B of~\citet{louizos2018learning} for more information about the density of the resulting distribution and its cumulative density function.

\section{Regularization} \label{app:regularization}
The regularization term is a KL divergence between the output distributions of the original model $p_{O}$ and the one from the model with interventions $p_{I}$ averaged at every token position: $\gL_{KL} =$
\begin{align}
    \frac{1}{T} \sum_{t=1}^T D_{KL}\left( p_O(x_t | x_{<t})~\|~p_{I}(x_t | x_{<t}) \right)
\end{align}
We sum $\gL_{KL}$ to Equation~\ref{eq:lagrangian_objective} multiplied by a factor. This factor is a hyperparameter that controls the amount of regularization to apply, and we empirically found that $1.0$ is a good value. In practice, as we discussed in Section~\ref{sec:experiments} the regularization term does not play an important role.

\section{Data} \label{app:data}

\paragraph{Subject-verb number agreement}
For this task, we employ data made available by~\citet{lakretz-etal-2019-emergence}.
The data are synthetic and generated with a modified version from~\citet{linzen-etal-2016-assessing} and~\citet{gulordava-etal-2018-colorless}.
Each synthetic number-agreement instance has a fixed syntax and varied lexical material.
Sentences were randomly sampled by choosing words from pools of 20 subject/object nouns, 15 verbs, 10 adverbs, 5 prepositions, 10 proper nouns and 10 location nouns. We used a total of 11000 training sentences and 1000 evaluation sentences.

\paragraph{Gender bias}
For this task, we employ data made available by~\citet{vig2020investigating}. The data are synthetic and generated with a list of templates from~\citet{lu2020gender} and several other templates, instantiated with professions from~\citet{bolukbasi2016man} (17 templates and 169 professions, resulting in 2,873 examples in total). We refer to~\citet{vig2020investigating} for the full lists of templates and professions. The templates have the form ``The [occupation] [verb] because [he/she]''. Professions are definitionally gender-neutral. We used a total of 2673 training sentences and 200 evaluation sentences.

\section{Additional results} \label{app:additional_results}

\begin{table}[t]
    \centering
    \begin{tabular}{rrrr}
    \toprule
    \textbf{Unit} & \textbf{Singular} & \textbf{Plural} & \textbf{Prevalence} \\
    \midrule
    79 & -0.93 \tiny{$\pm 0.03$}  & 0.96 \tiny{$\pm 0.02$} & 100\% \\
    93 & 0.94 \tiny{$\pm 0.02$} & -0.81 \tiny{$\pm 0.05$} & 100\% \\
    357 & -0.97 \tiny{$\pm 0.01$} & 0.83 \tiny{$\pm 0.06$} & 30\% \\
    498 & 0.96 \tiny{$\pm 0.01$} & -0.92 \tiny{$\pm 0.02$} & 100\% \\
    571 & -0.96 \tiny{$\pm 0.02$} & 0.87 \tiny{$\pm 0.04$} & 60\% \\
    630 & 0.91 \tiny{$\pm 0.03$} & -0.34 \tiny{$\pm 0.19$} & 100\% \\
    \bottomrule
    \end{tabular}
    \caption{Subject-verb number agreement task with single-step interventions. Values are averages across 10 runs. }
    \label{tab:svna_single}
\end{table}

\begin{table}[t]
    \centering
    \begin{tabular}{rrrr}
    \toprule
    \textbf{Unit} & \textbf{He} & \textbf{She} & \textbf{Prevalence} \\
    \midrule
    5 & 1.00 \tiny{$\pm 0.00$}  & -0.99 \tiny{$\pm 0.01$} & 50\% \\
    288 & -1.00 \tiny{$\pm 0.00$}  & -0.89 \tiny{$\pm 0.21$} & 100\% \\
    336 & -1.00 \tiny{$\pm 0.00$}  & 1.00 \tiny{$\pm 0.00$} & 20\% \\
    455 & -1.00 \tiny{$\pm 0.00$}  & -0.24 \tiny{$\pm 0.00$} & 10\% \\
    456 & 1.00 \tiny{$\pm 0.00$}  & -1.00 \tiny{$\pm 0.00$} & 100\% \\
    464 & 1.00 \tiny{$\pm 0.00$}  & -1.00 \tiny{$\pm 0.00$} & 20\% \\
    474 & 1.00 \tiny{$\pm 0.00$}  & -0.72 \tiny{$\pm 0.18$} & 60\% \\
    490 & -1.00 \tiny{$\pm 0.00$}  & 1.00 \tiny{$\pm 0.00$} & 20\% \\
    563 & -1.00 \tiny{$\pm 0.00$}  & 1.00 \tiny{$\pm 0.00$} & 100\% \\
    646 & -1.00 \tiny{$\pm 0.00$}  & 1.00 \tiny{$\pm 0.00$} & 10\% \\
    \bottomrule
    \end{tabular}
    \caption{Gender bias task with single-step interventions. Values are averages across 10 runs.}
    \label{tab:gb_single}
\end{table}

\begin{table}[t]
    \centering
    \begin{tabular}{rrrr}
    \toprule
    \textbf{Unit} & \textbf{He} & \textbf{She} & \textbf{Prevalence} \\
    \midrule
    456 & 0.91 \tiny{$\pm 0.28$}  & -0.94 \tiny{$\pm 0.17$} & 100\% \\
    670 & 0.86 \tiny{$\pm 0.27$}  & -0.55 \tiny{$\pm 0.35$} & 80\% \\
    693 & 1.00 \tiny{$\pm 0.00$}  & -1.00 \tiny{$\pm 0.00$} & 90\% \\
    1009 & 0.49 \tiny{$\pm 0.00$}  & -0.21 \tiny{$\pm 0.00$} & 10\% \\
    1184 & -1.00 \tiny{$\pm 0.00$}  & 1.00 \tiny{$\pm 0.00$} & 100\% \\
    1252 & 0.79 \tiny{$\pm 0.27$}  & -1.00 \tiny{$\pm 0.00$} & 50\% \\
    \bottomrule
    \end{tabular}
    \caption{Gender bias task with every-step interventions. Values are averages across 10 runs.}
    \label{tab:gb_every}
\end{table}

\begin{figure*}[t]
    \centering
    \begin{subfigure}[b]{0.48\textwidth}
        \centering
        \includegraphics[width=\textwidth]{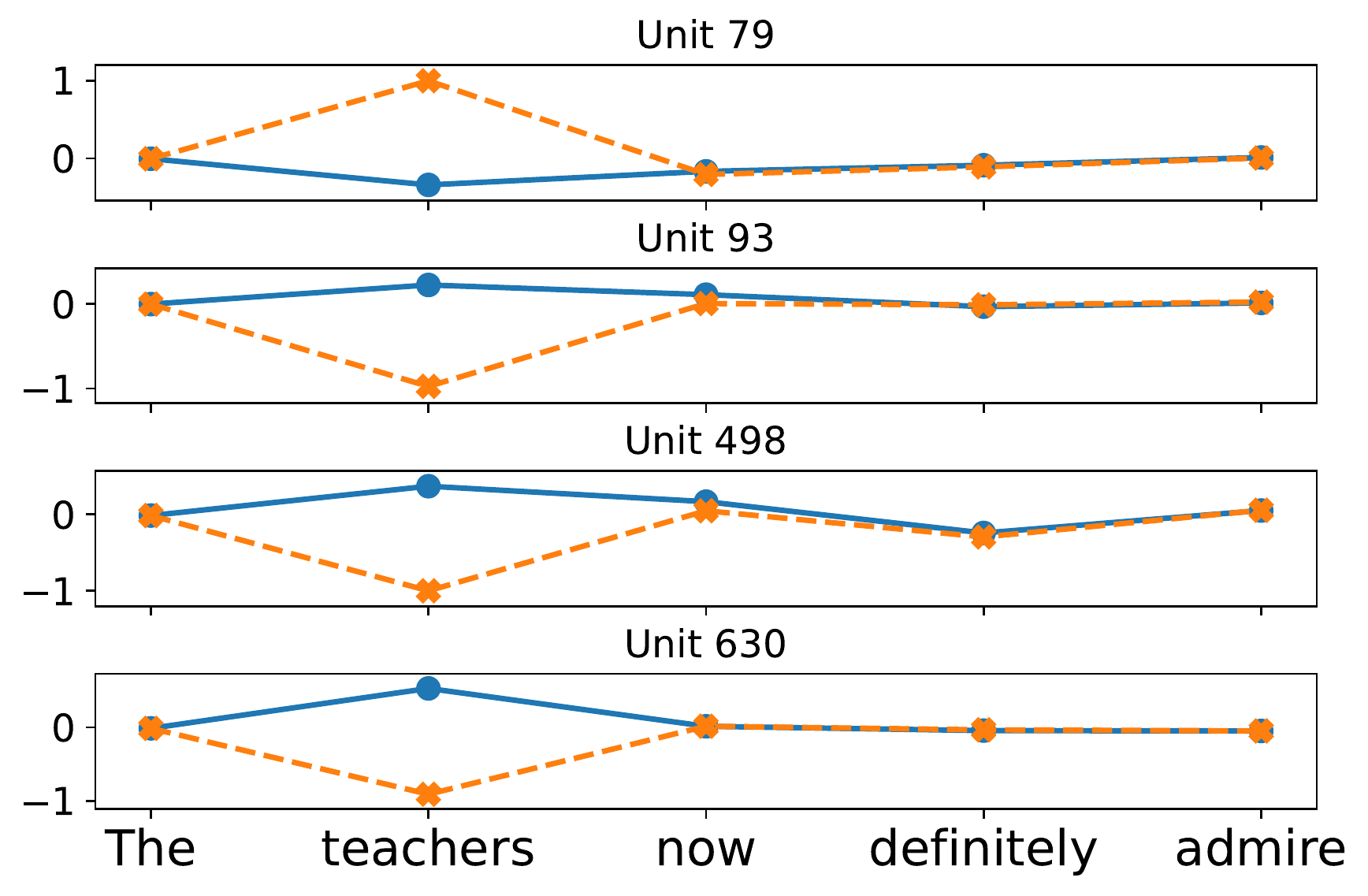}
        \caption{}
    \end{subfigure}
    \hfill
    \begin{subfigure}[b]{0.48\textwidth}
        \centering
        \includegraphics[width=\textwidth]{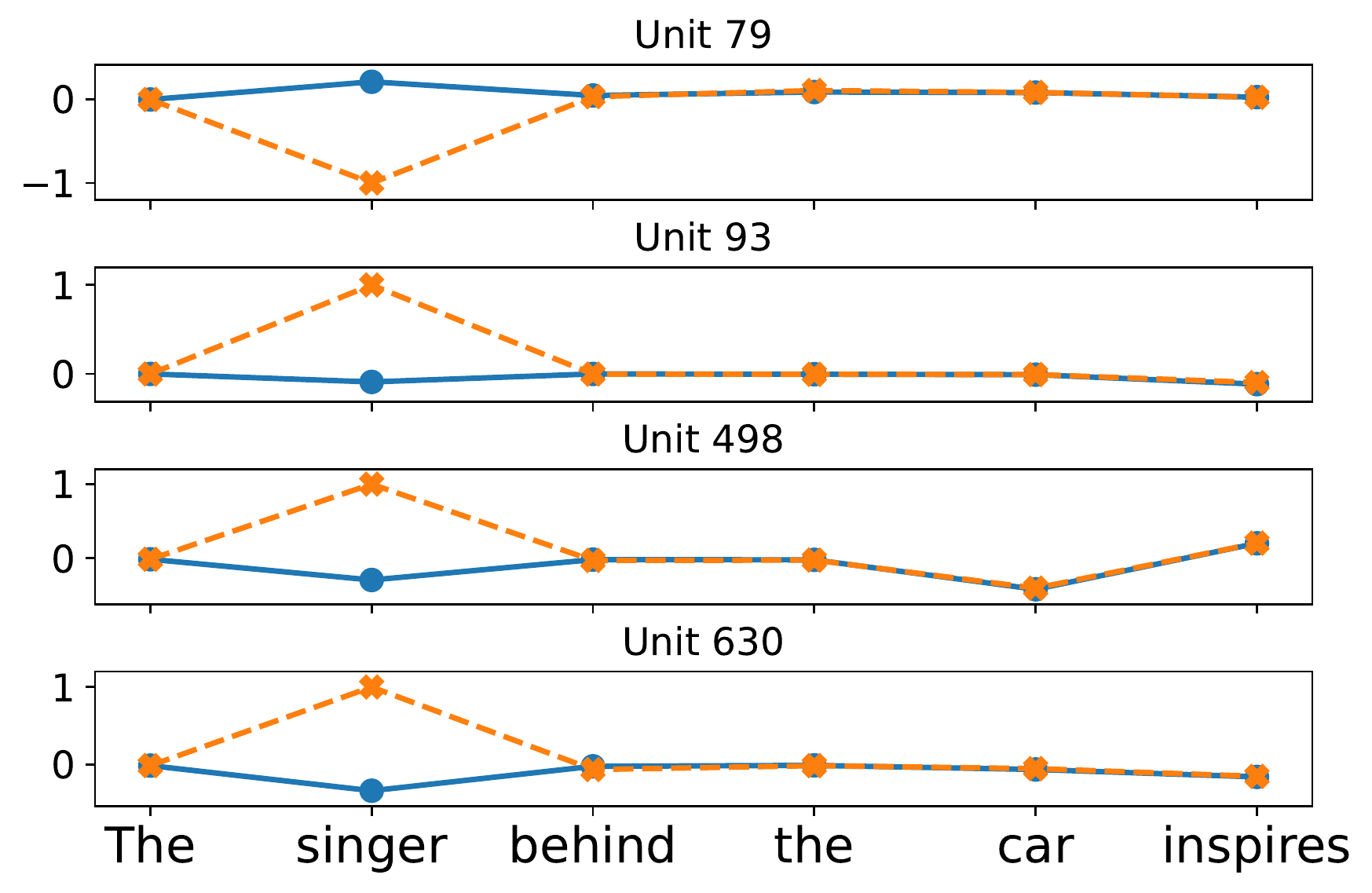}
        \caption{}
    \end{subfigure}
    \par\vspace{0.5em}
    \begin{subfigure}[b]{0.48\textwidth}
        \centering
        \includegraphics[width=\textwidth]{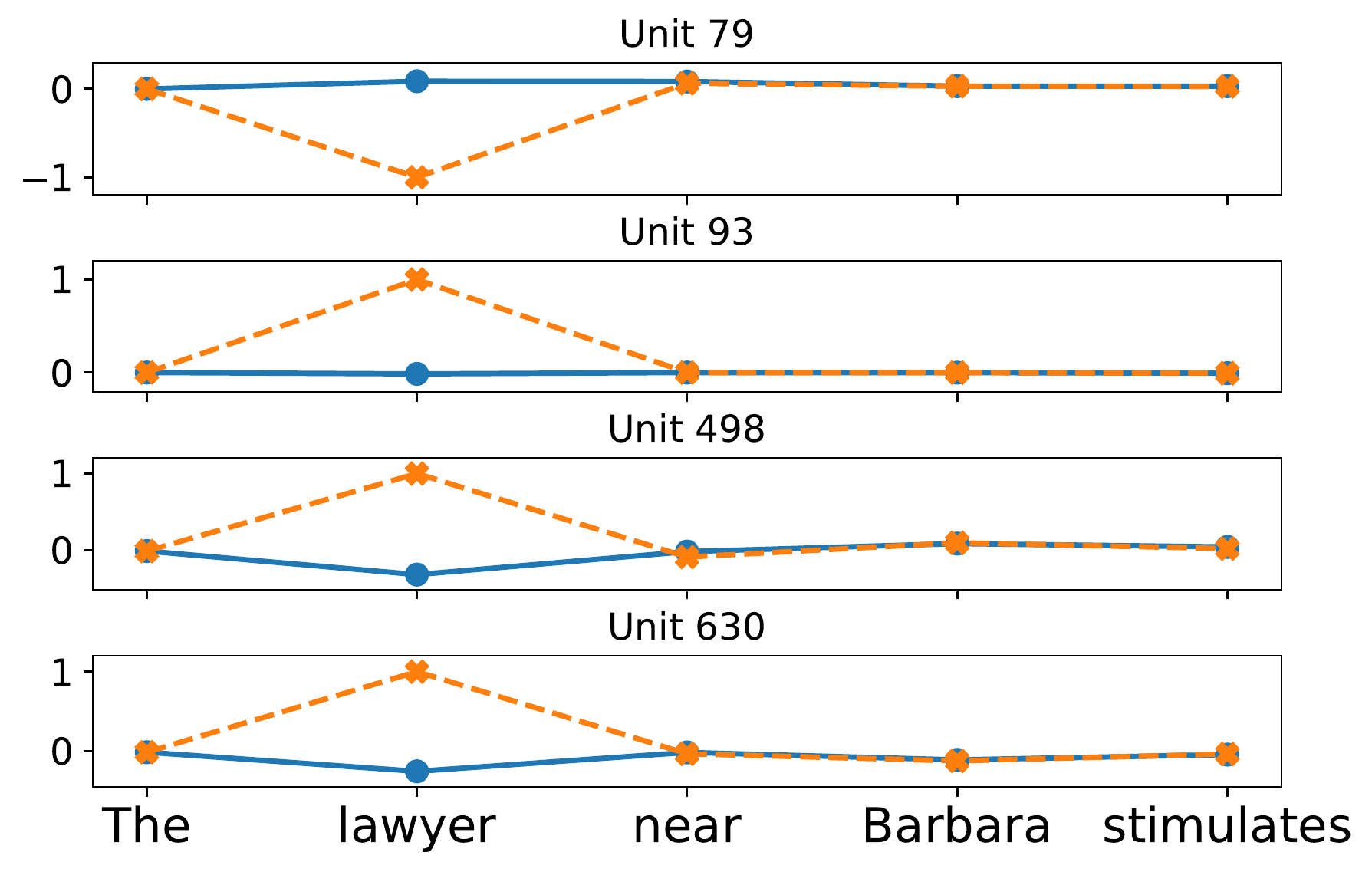}
        \caption{}
    \end{subfigure}
    \hfill
    \begin{subfigure}[b]{0.48\textwidth}
        \centering
        \includegraphics[width=\textwidth]{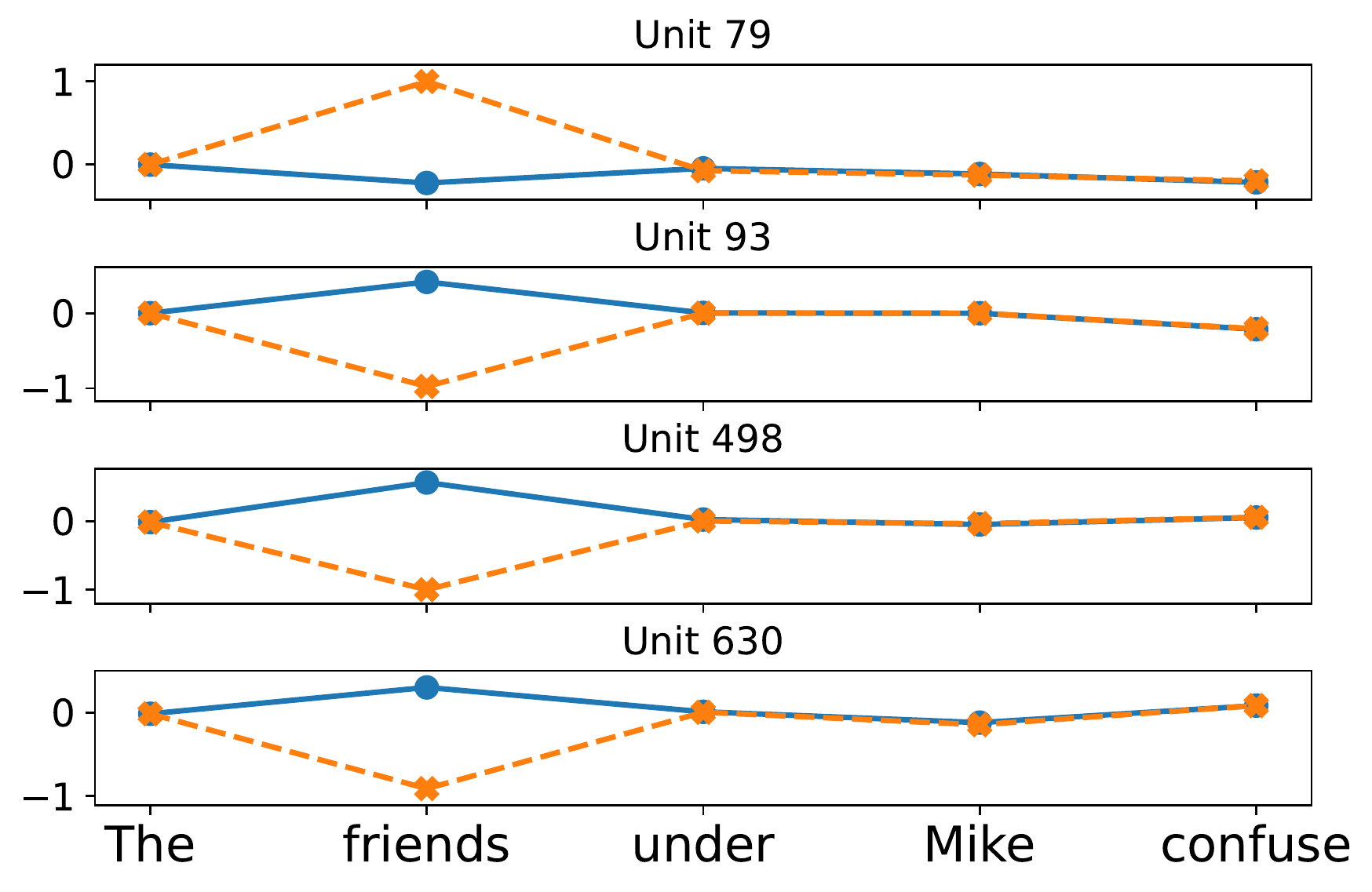}
        \caption{}
    \end{subfigure}
    \par\vspace{0.5em}
    \begin{subfigure}[b]{0.48\textwidth}
        \centering
        \includegraphics[width=\textwidth]{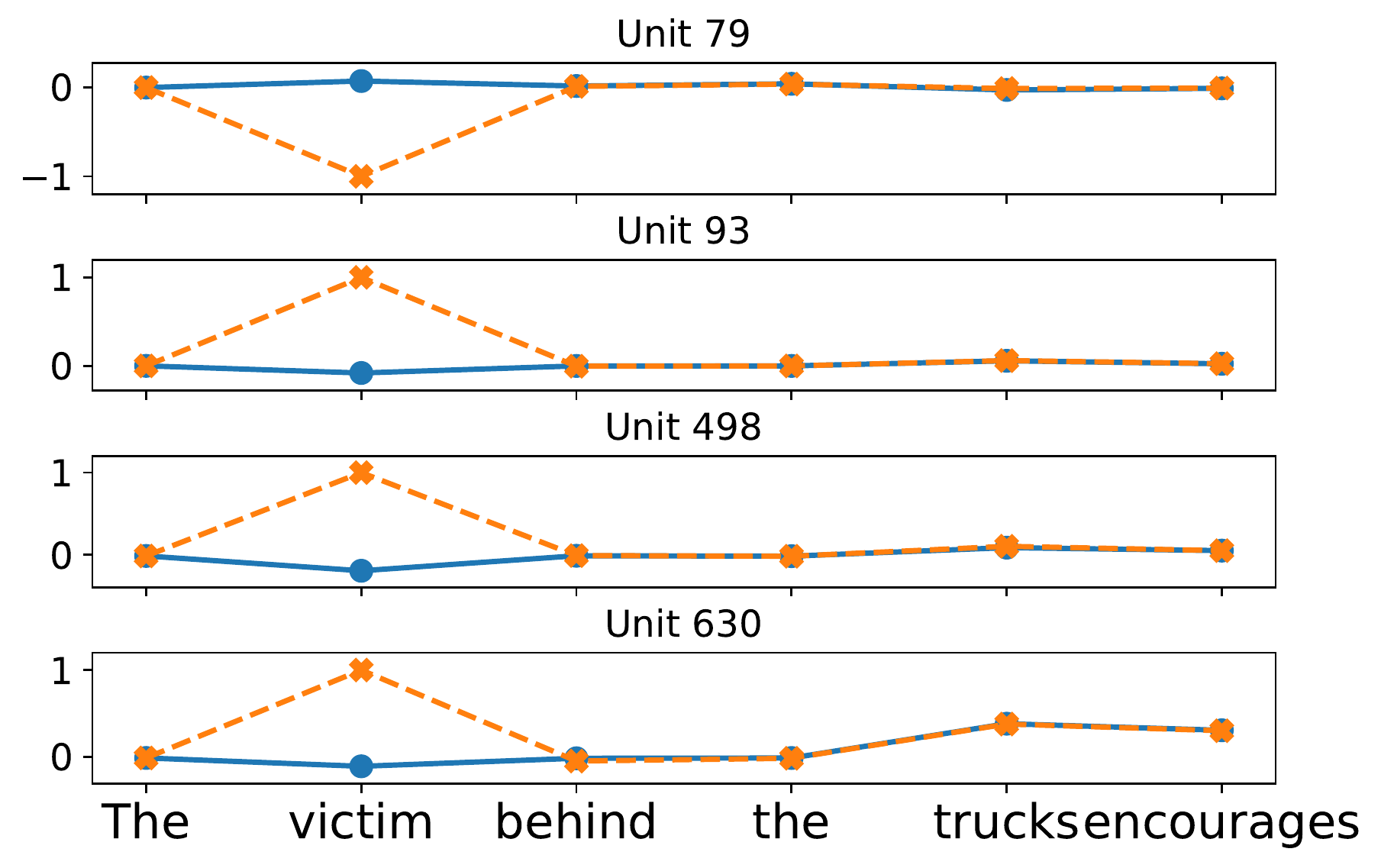}
        \caption{}
    \end{subfigure}
    \hfill
    \begin{subfigure}[b]{0.48\textwidth}
        \centering
        \includegraphics[width=\textwidth]{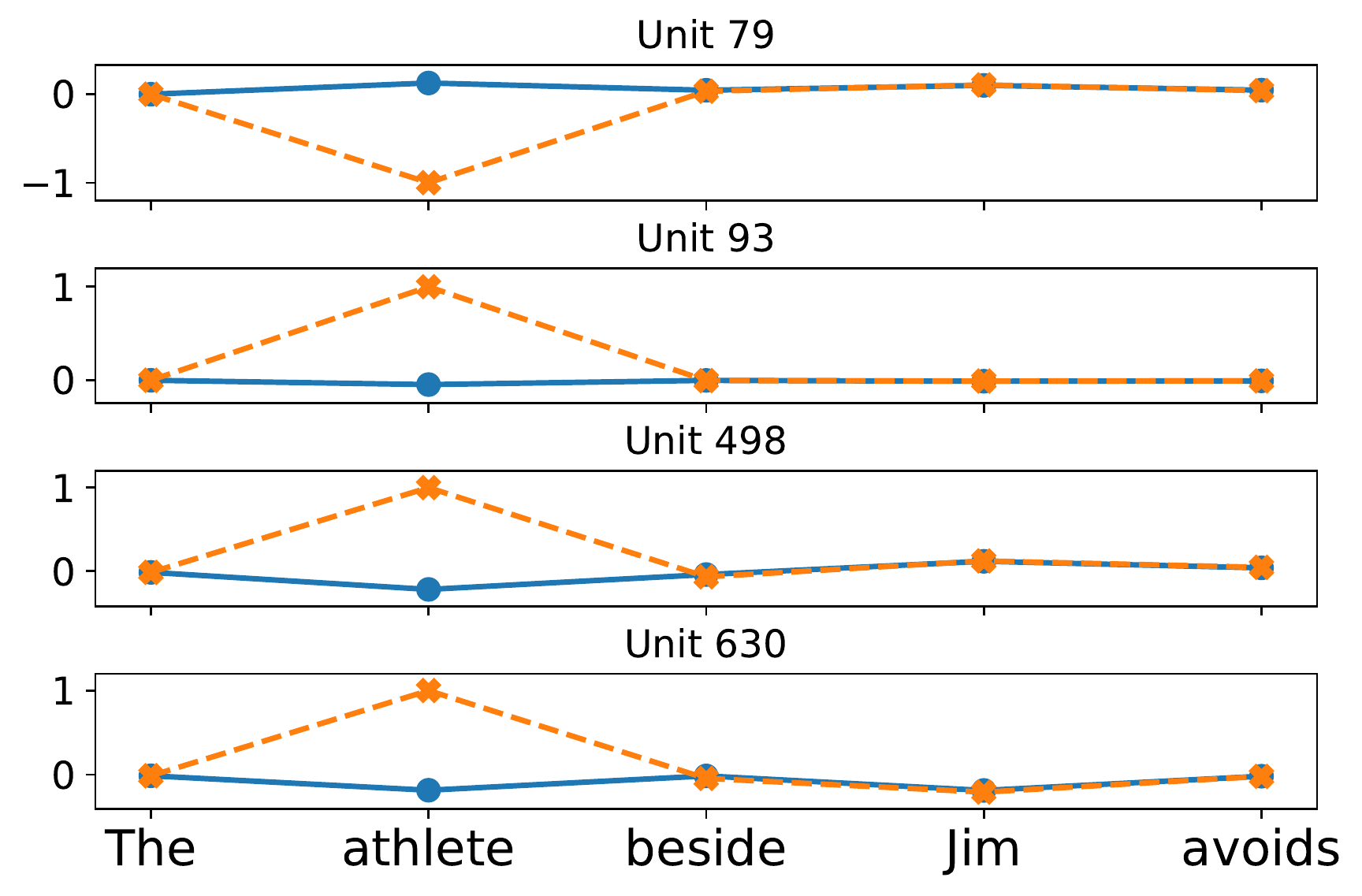}
        \caption{}
    \end{subfigure}
    \par\vspace{0.5em}
    \begin{subfigure}[b]{0.48\textwidth}
        \centering
        \includegraphics[width=\textwidth]{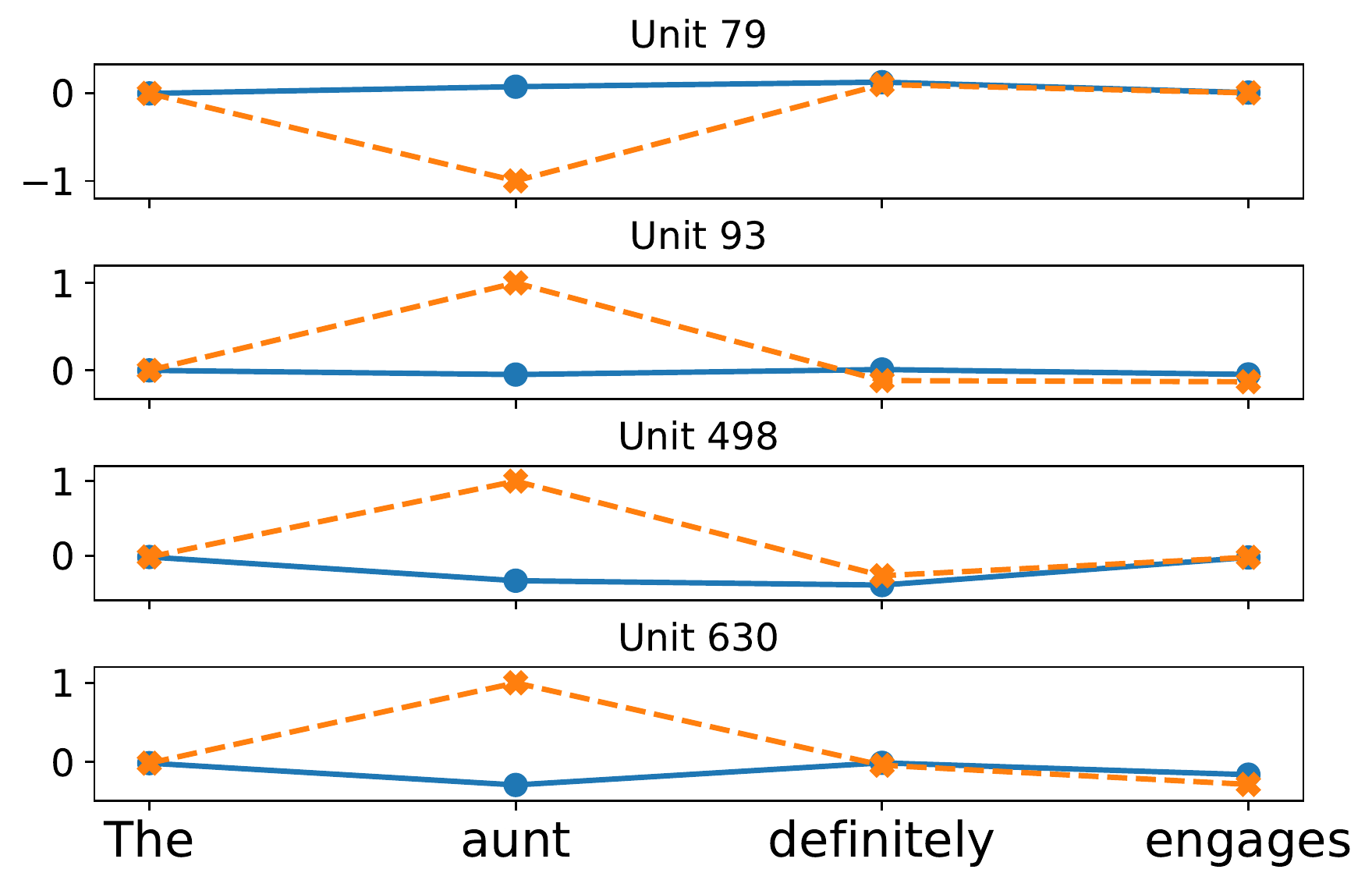}
        \caption{}
    \end{subfigure}
    \hfill
    \begin{subfigure}[b]{0.48\textwidth}
        \centering
        \includegraphics[width=\textwidth]{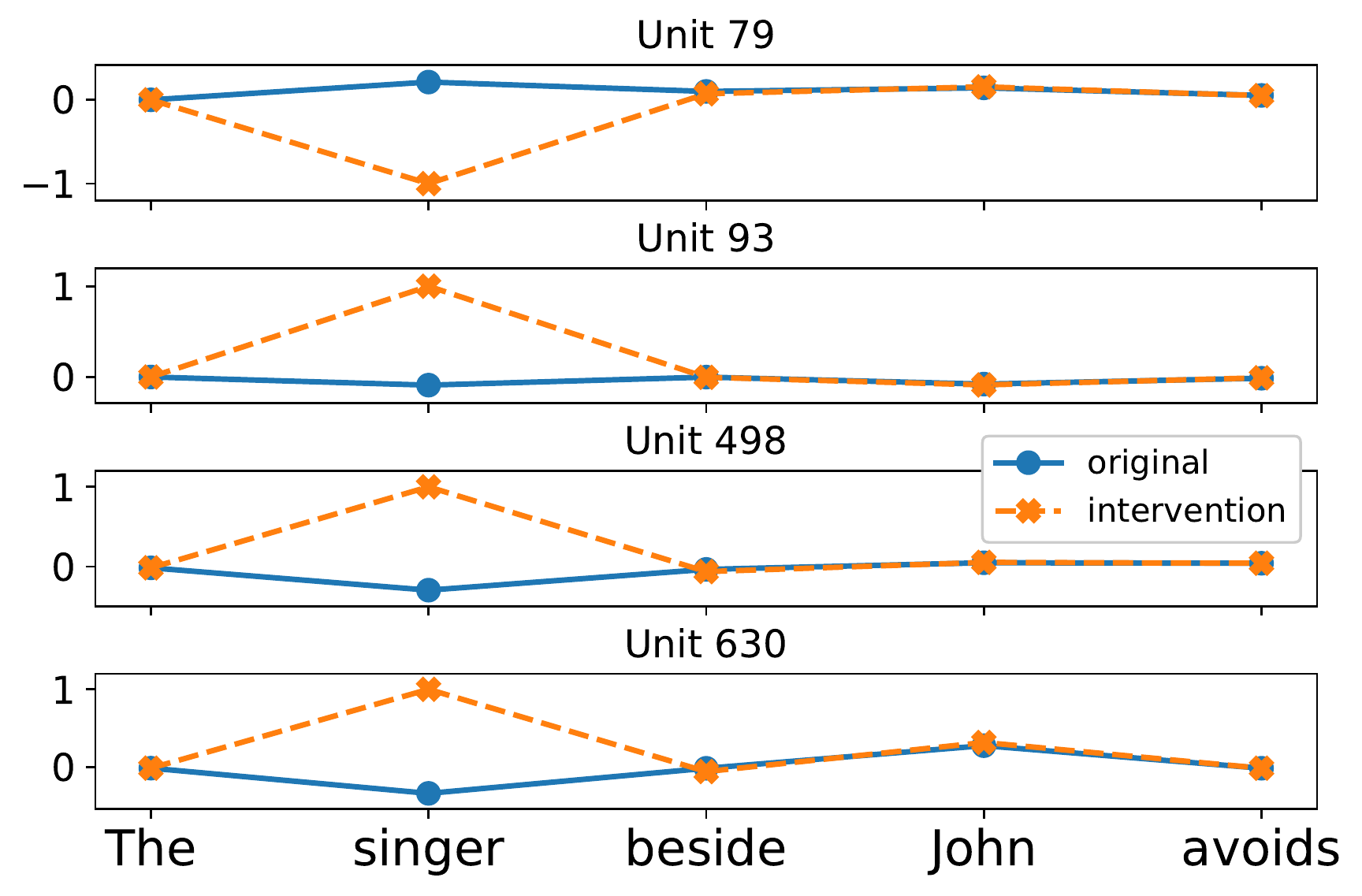}
        \caption{}
    \end{subfigure}
    \caption{Subject-verb number agreement: activations of four units we intervene on (single step intervention at the second token from the left) for changing number agreement (at the last token).}
    \label{fig:extra_example_na}
\end{figure*}

\begin{figure*}[t]
    \centering
    \begin{subfigure}[b]{0.48\textwidth}
        \centering
        \includegraphics[width=\textwidth]{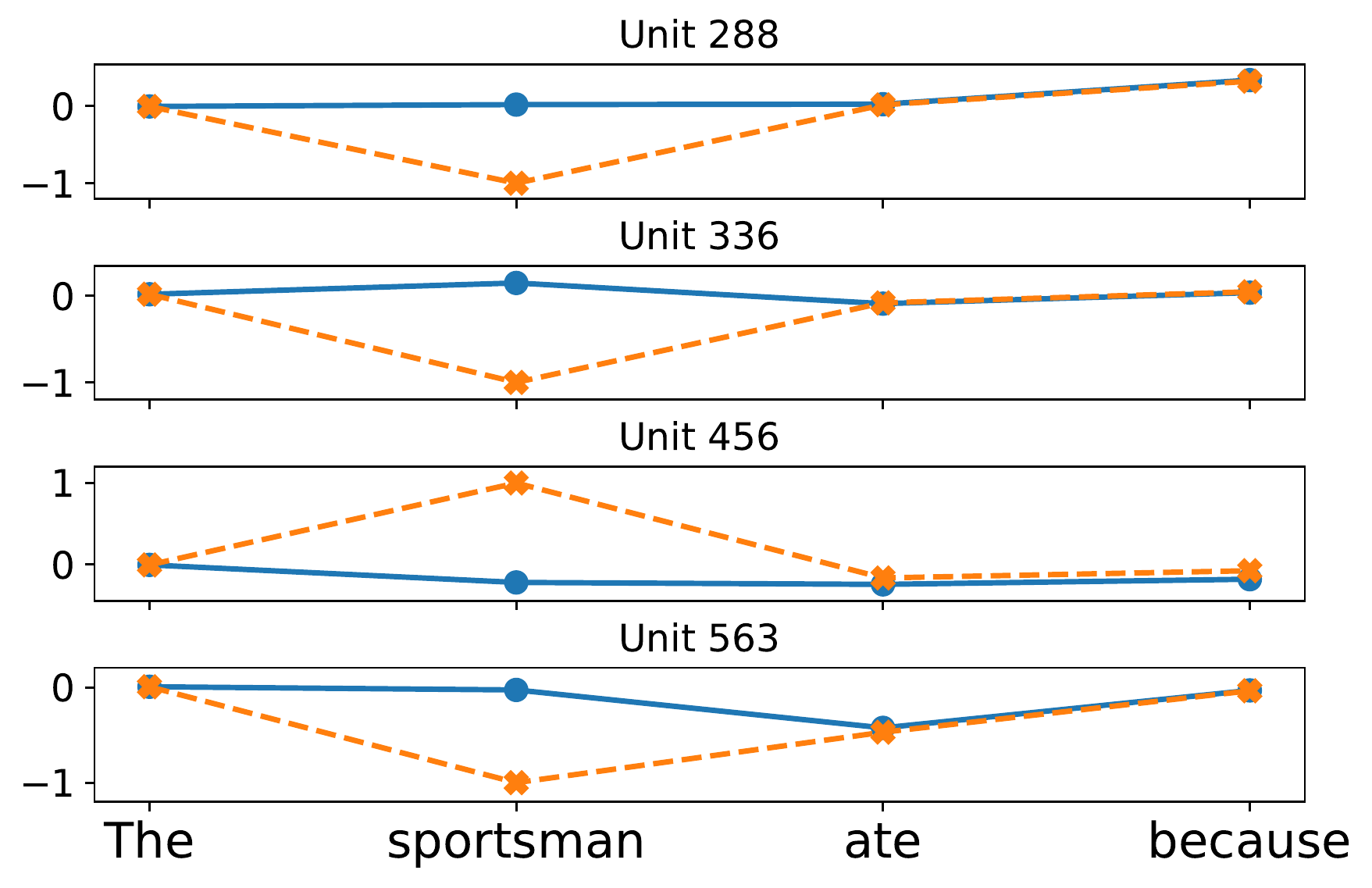}
        \caption{}
    \end{subfigure}
    \hfill
    \begin{subfigure}[b]{0.48\textwidth}
        \centering
        \includegraphics[width=\textwidth]{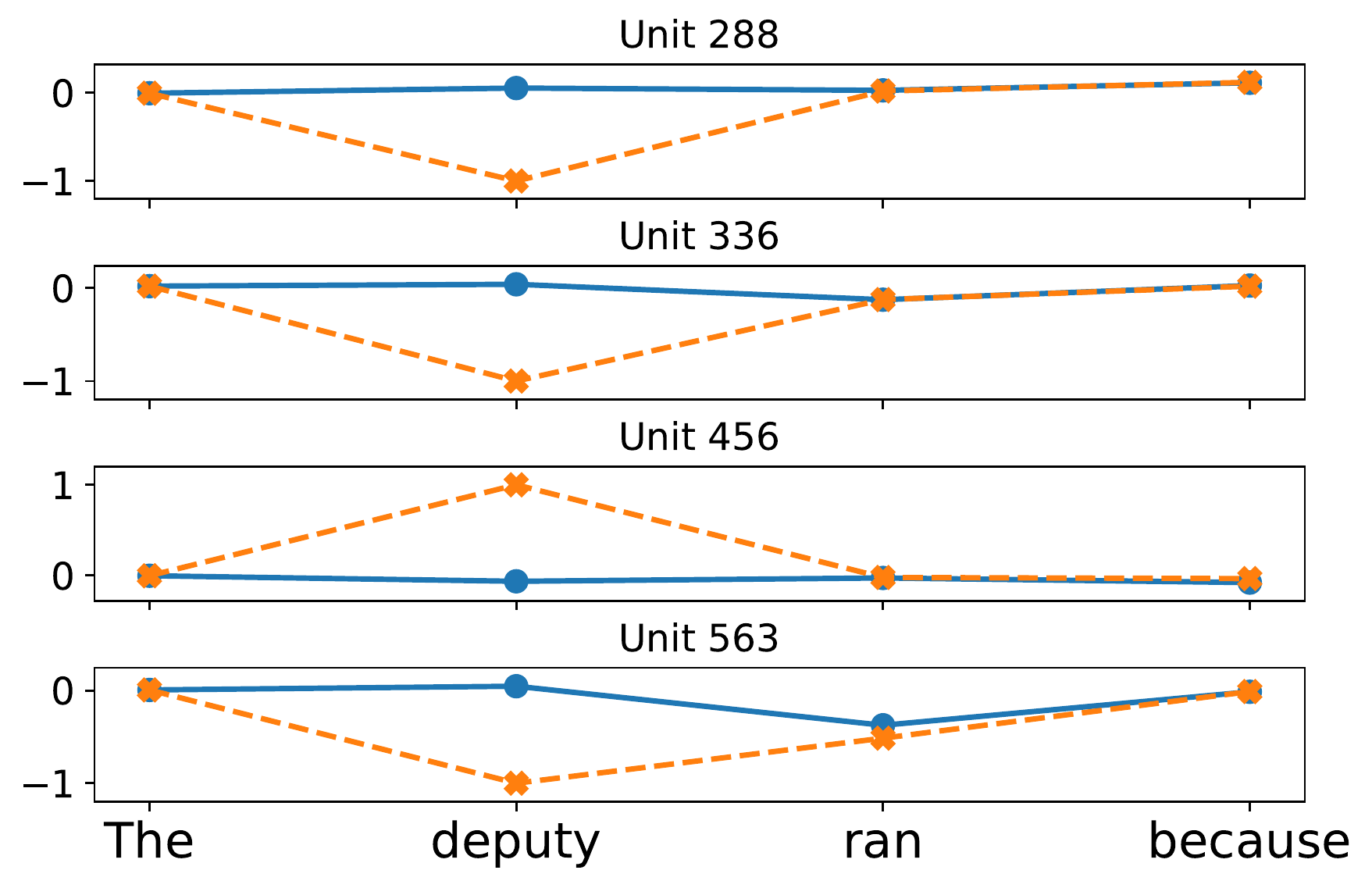}
        \caption{}
    \end{subfigure}
    \par\vspace{0.5em}
    \begin{subfigure}[b]{0.48\textwidth}
        \centering
        \includegraphics[width=\textwidth]{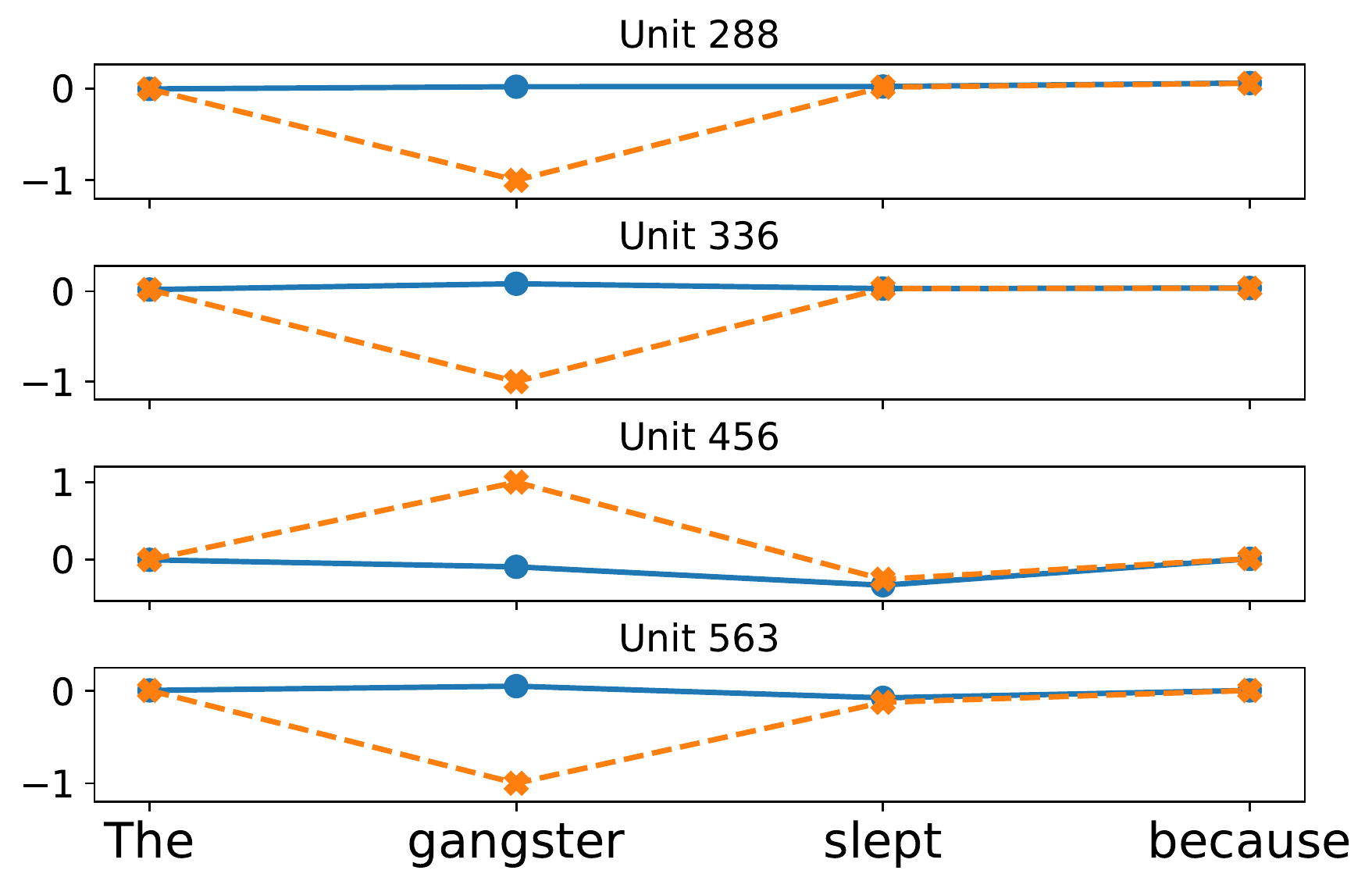}
        \caption{}
    \end{subfigure}
    \hfill
    \begin{subfigure}[b]{0.48\textwidth}
        \centering
        \includegraphics[width=\textwidth]{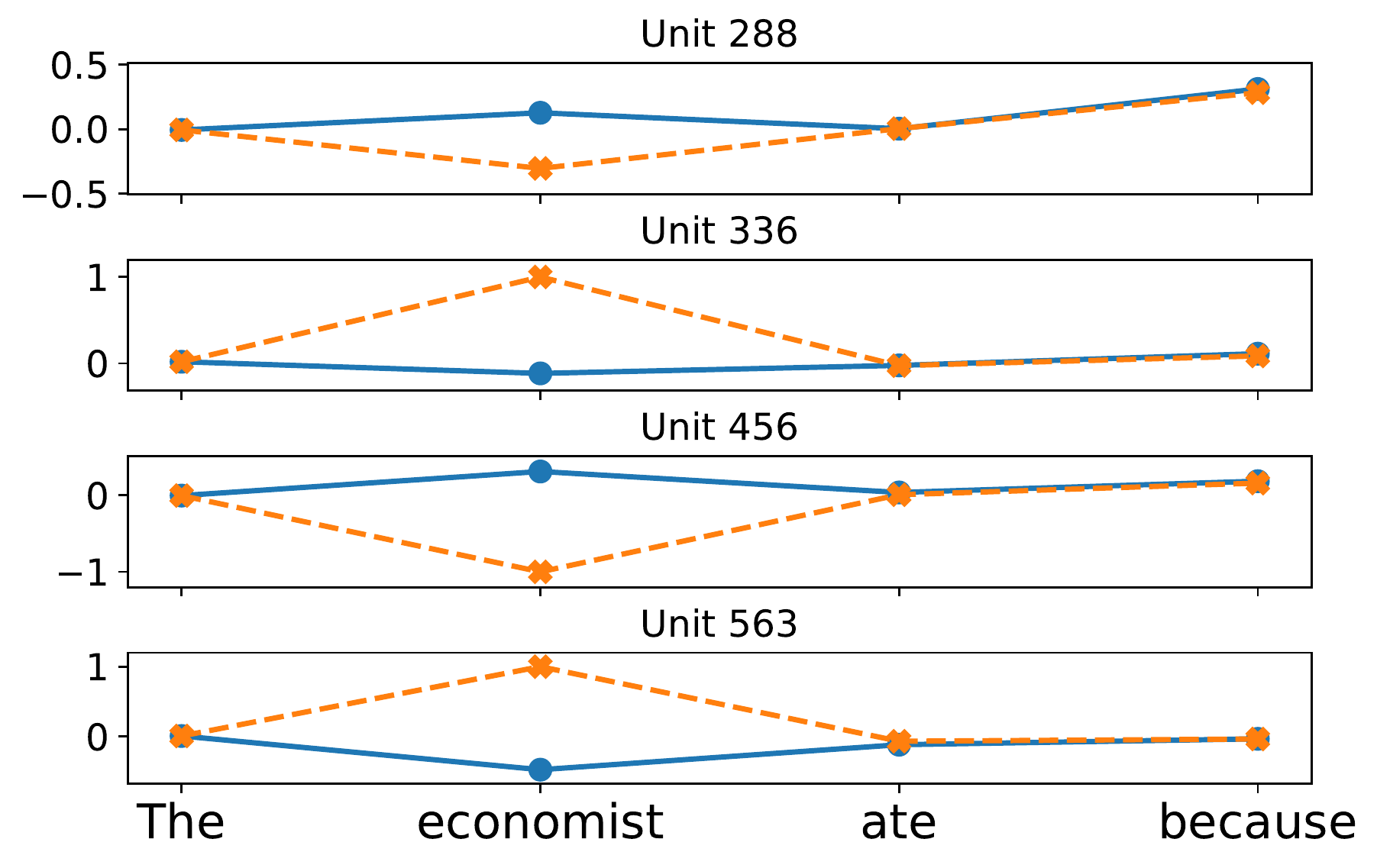}
        \caption{}
    \end{subfigure}
    \par\vspace{0.5em}
    \begin{subfigure}[b]{0.48\textwidth}
        \centering
        \includegraphics[width=\textwidth]{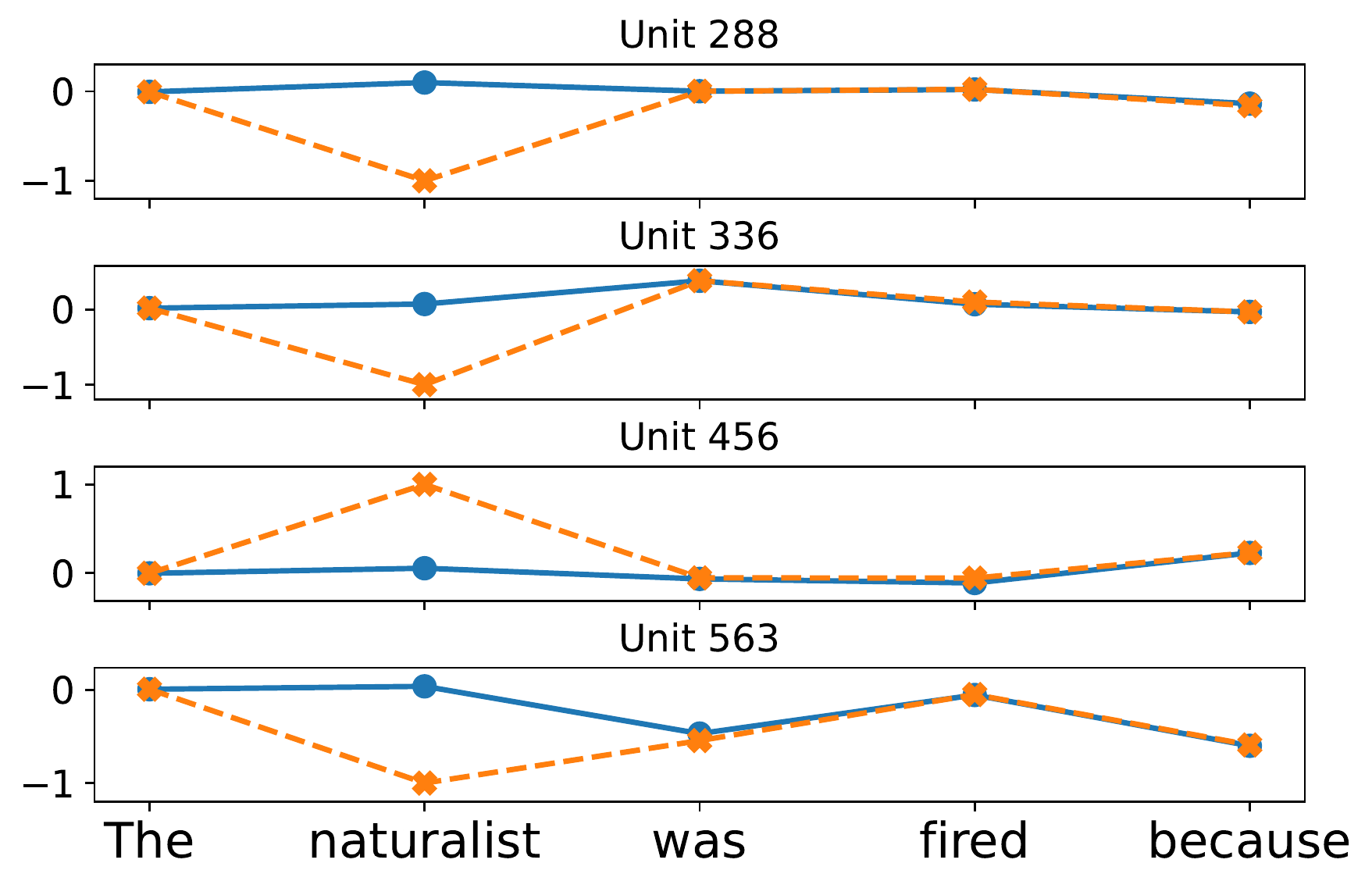}
        \caption{}
    \end{subfigure}
    \hfill
    \begin{subfigure}[b]{0.48\textwidth}
        \centering
        \includegraphics[width=\textwidth]{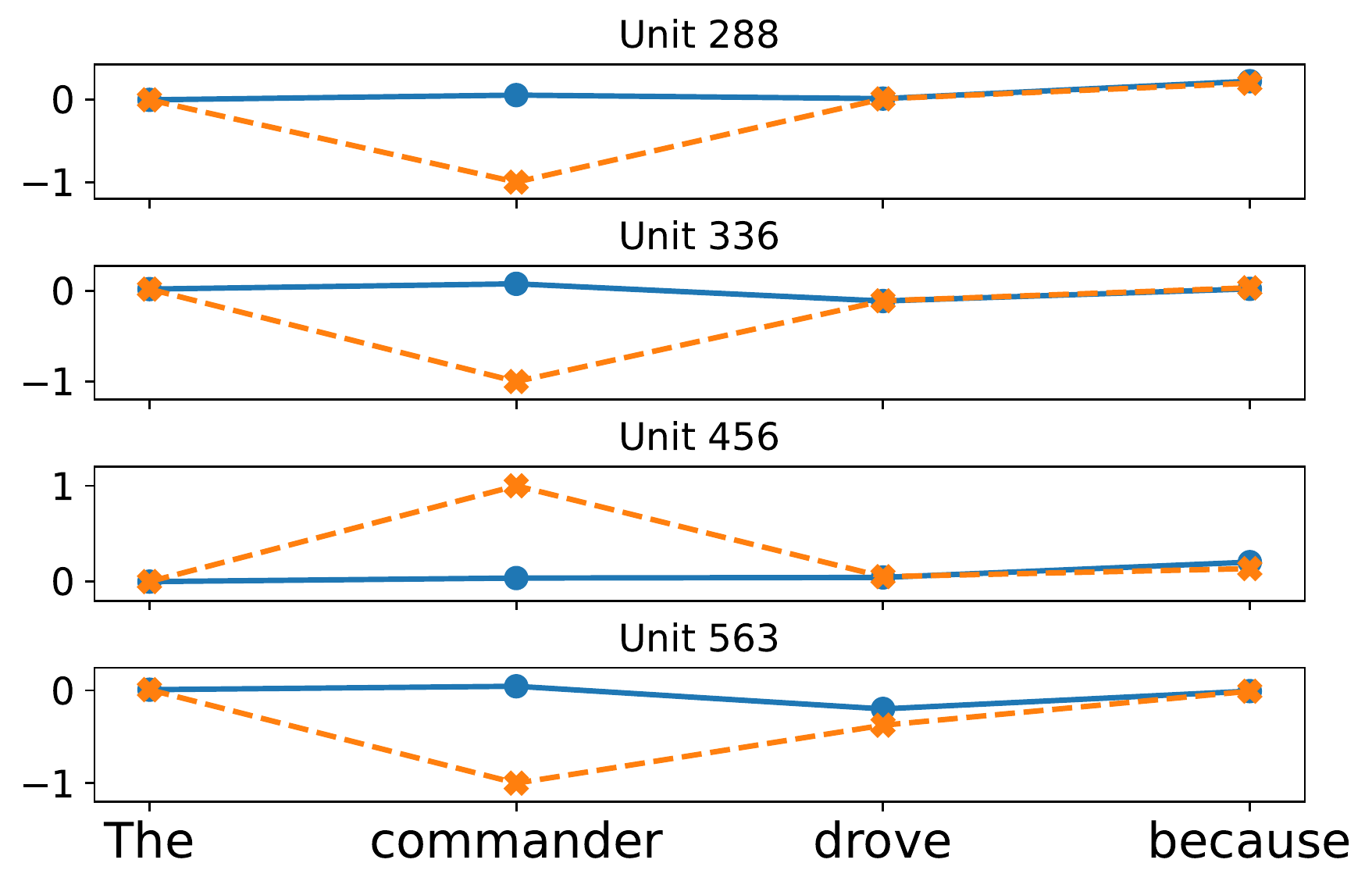}
        \caption{}
    \end{subfigure}
    \par\vspace{0.5em}
    \begin{subfigure}[b]{0.48\textwidth}
        \centering
        \includegraphics[width=\textwidth]{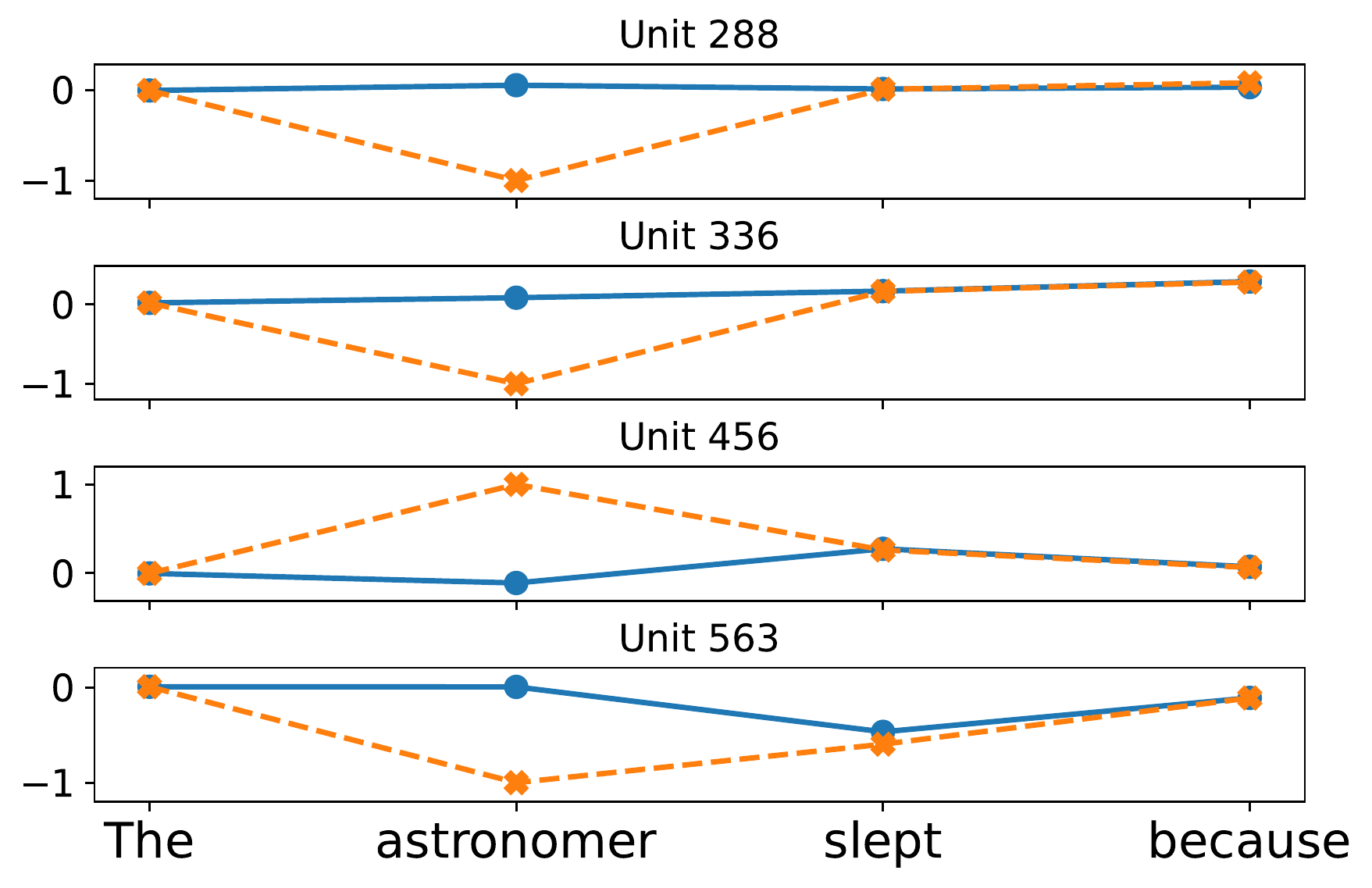}
        \caption{}
    \end{subfigure}
    \hfill
    \begin{subfigure}[b]{0.48\textwidth}
        \centering
        \includegraphics[width=\textwidth]{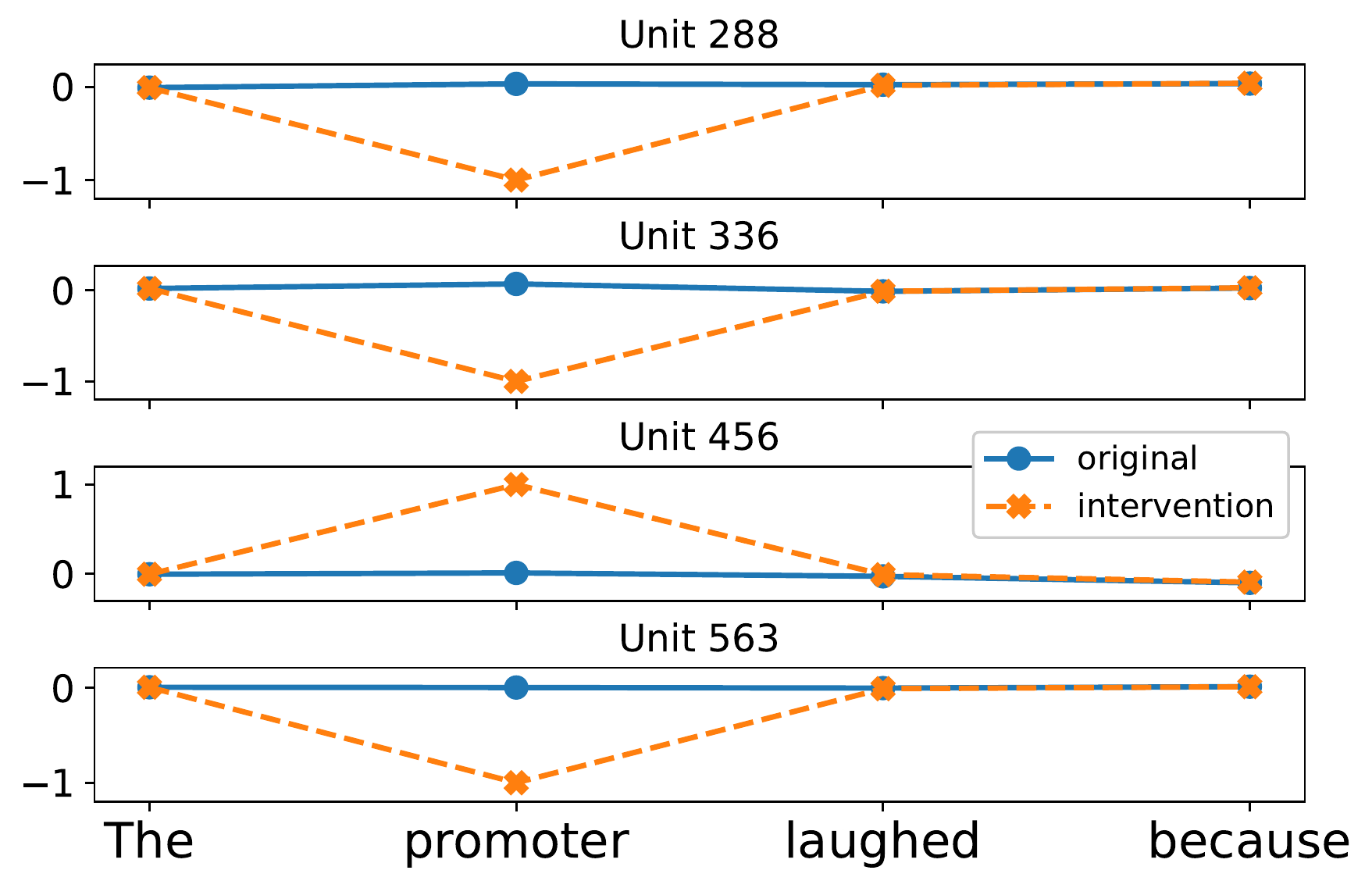}
        \caption{}
    \end{subfigure}
    \caption{Gender bias: activations of four units we intervene on (single step intervention at the second token from the left) for changing the pronoun (after ``because'').}
    \label{fig:extra_example_gb}
\end{figure*}

\end{document}